\documentclass[lettersize,journal]{IEEEtran}
\usepackage{amsmath,amsfonts}
\usepackage{algorithmic}
\usepackage{algorithm}
\usepackage{array}
\usepackage[caption=false,font=normalsize,labelfont=sf,textfont=sf]{subfig}
\usepackage{textcomp}
\usepackage{stfloats}
\usepackage{url}
\usepackage{verbatim}
\usepackage{graphicx}
\usepackage{cite}
\usepackage{url}
\usepackage{amsfonts}
\usepackage[graphicx]{realboxes}
\usepackage{array}
\usepackage{booktabs}
\usepackage{tcolorbox}  
\usepackage{esvect}
\usepackage{tabularx}  
\usepackage{ragged2e}
\usepackage{multirow}
\usepackage{caption}
\usepackage{colortbl}
\usepackage{wrapfig}
\usepackage{enumitem}
\usepackage{arydshln}
\usepackage{float}
\usepackage{makecell}
\usepackage[normalem]{ulem}
\newcounter{mycounter}

\begin{document}

\title{Atomic Fact Decomposition Helps Attributed Question Answering}
\author{Zhichao Yan, Jiapu Wang, Jiaoyan Chen, Xiaoli Li, \IEEEmembership{Fellow, IEEE}, Jiye Liang, \IEEEmembership{Fellow, IEEE}, 

Ru Li, \IEEEmembership{Member, IEEE}, Jeff Z.Pan, \IEEEmembership{Member, IEEE}
\thanks{Zhichao Yan, Jiye Liang and Ru Li are with the School of Computer and Information Technology, Shanxi University, Taiyuan, Shanxi 030006, China (e-mail: 202312407023@email.sxu.edu.cn; ljy@sxu.edu.cn; liru@sxu.edu.cn)}
\thanks{Jiapu Wang is with the Beijing Key Laboratory of Multimedia and Intelligent Software Technology, Beijing Institute of Artificial Intelligence, School of Information Science and Technology, Beijing University of Technology, Beijing 100124, China (e-mail: jpwang@emails.bjut.edu.cn)}
\thanks{Jiaoyan Chen is with the University of Manchester (e-mail: jiaoyan.chen@manchester.ac.uk)}
\thanks{Xiaoli Li is with the Singapore University of Technology and Design, Singapore (e-mail: xiaoli\_li@sutd.edu.sg)}
\thanks{Jeff Z. Pan is with ILCC, School of Informatics, University of Edinburgh, Edinburgh, UK (e-mail: j.z.pan@ed.ac.uk)}
\thanks{Manuscript received XX, 20XX; revised XX, 20XX. (corresponding author: Ru Li and Jeff Z.Pan)}
}


\markboth{Journal of \LaTeX\ Class Files,~Vol.~14, No.~8, August~2021}%
{Shell \MakeLowercase{\textit{et al.}}: A Sample Article Using IEEEtran.cls for IEEE Journals}


\maketitle

\begin{abstract}
  Attributed Question Answering (AQA) aims to provide both a trustworthy answer and a reliable attribution report for a given question. 
  Retrieval is a widely adopted approach, including two general paradigms: Retrieval-Then-Read (RTR) and post-hoc retrieval.
  Recently, Large Language Models (LLMs) have shown remarkable proficiency, prompting growing interest in AQA among researchers.  
  However, RTR-based AQA often suffers from irrelevant knowledge and rapidly changing information, even when LLMs are adopted, while post-hoc retrieval-based AQA struggles with comprehending long-form answers with complex logic, and precisely identifying the content needing revision and preserving the original intent.
  To tackle these problems, this paper proposes an \textbf{A}tomic fact decomposition-based \textbf{R}etrieval and \textbf{E}diting \textbf{(ARE)} framework, which decomposes the generated long-form answers into molecular clauses and atomic facts by the instruction-tuned LLMs. Notably, the instruction-tuned LLMs are fine-tuned using a well-constructed dataset, generated from large scale Knowledge Graphs (KGs). This process involves extracting one-hop neighbors from a given set of entities and transforming the result into coherent long-form text. Subsequently, ARE leverages a search engine to retrieve evidences related to atomic facts, inputting these evidences into an LLM-based verifier to determine whether the facts require expansion for re-retrieval or editing.
  Furthermore, the edited facts are backtracked into the original answer, with evidence aggregated based on the relationship between molecular clauses and atomic facts. Extensive evaluations demonstrate the superior performance of our proposed method over the state-of-the-arts on several datasets, with an additionally proposed new metric $Attr_{p}$ for evaluating the precision of evidence attribution. 
\end{abstract}

\begin{IEEEkeywords}
Attributed Question Answer, Information Retrieval, Large Language Models.
\end{IEEEkeywords}

\section{Introduction}
\IEEEPARstart{L}{arge} Language Models (LLMs), pre-trained on large-scale text corpora \cite{10387715}, have demonstrated remarkable capabilities in natural language understanding and generation tasks \cite{zhuang2023toolqa,wang2024large}. However, their application in real-world scenarios is hindered by a lack of transparency, limited interpretability, and occasional factual inaccuracies,  \cite{huang2023survey,peskoff-stewart-2023-credible}.These limitations compromise the reliability of LLM outputs, especially in settings where factual correctness and evidence traceability are critical.

To address these challenges, Attributed Question Answering (AQA) \cite{li2023survey} has emerged as a promising paradigm. AQA aims to enhance answer trustworthiness by explicitly linking generated responses to verifiable sources of evidence. As illustrated in Figure 1, when asked “Who has the most Super Bowl rings?”, an LLM initially responds: “The player with the most Super Bowl rings is Tom Brady. Tom Brady is an American football quarterback who has won six Super Bowl championships.” Upon retrieving external evidence, however, the system identifies a factual error — Tom Brady has won seven, not six, championships. This triggers a post-hoc correction process, resulting in a revised and properly attributed answer, along with the supporting evidence.

Depending on the timing of retrieval, previous research can be divided into two categories: Retrieval-then-Read (RTR) ~\cite{gao2023enabling}, and Post-hoc Retrieval \cite{bohnet2022attributed,gao2023rarr,chen2023purr,yan2025decomposing}. RTR delivers relevant answers by retrieving documents based on the query, providing detailed context that can enhance the richness of the response. 
However, LLMs often link content to irrelevant or incorrect sources and lack an effective mechanism for subsequent correction.
In contrast, post-hoc retrieval typically retrieves specific external information for initially generated long-form answers, effectively reducing linking errors and enabling targeted revisions, which enhances flexibility and robustness \cite{dhuliawala2023chainofverification,kim2024re}. 

\begin{figure*}[t]
    \centering
    \includegraphics[scale=0.12]{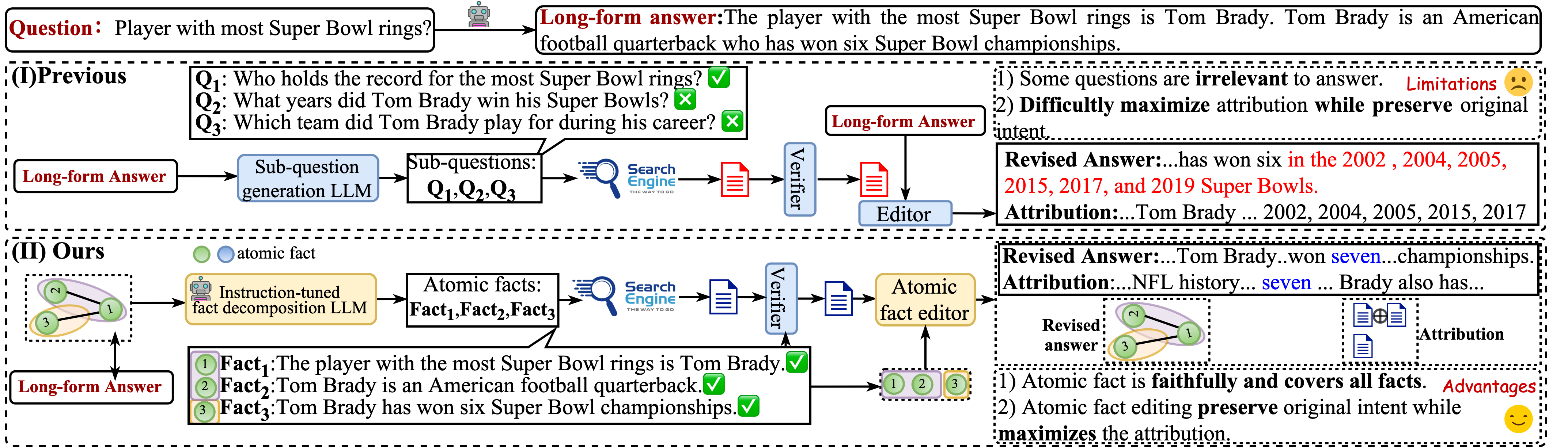}
    \caption{The motivation of previous methods and our proposed ARE. Previous methods directly use LLMs to generate sub-questions for the long-form answer and edit the whole answer. Our proposed ARE leverages instruction-tuned LLMs to achieve molecular-to-atomic stage, and backtrack the edited atomic facts to achieve atomic-to-molecular stage. 
    }
    \label{fig:introduction}
\end{figure*}

\setcounter{mycounter}{1}
Despite the flexibility post-hoc retrieval offers in verifying each fact and correcting it in long-form answers, it introduces several significant challenges. Firstly, existing methods~\cite{chen2023purr,kim2024re} prompt LLMs to directly generate a series of sub-questions for retrieval based on the long-form answer (As shown in Figure \ref{fig:introduction} (\Roman{mycounter})). However, the complex logical structures and vast amounts of information typically found in long-form answers make it difficult for these methods to capture all necessary facts comprehensively. As a result, the generated sub-questions often fail to align with the core content of the answer, leading to irrelevant questions and ultimately hindering the retrieval of accurate supporting evidence.

Moreover, existing methods \cite{gao2023rarr, wu2024updating} that directly edit long-form answers holistically often face difficulty in precisely locating specific content for revision, making it challenging to preserve the original intent while ensuring consistency between the answer and the attribution.  This limitation arises from either insufficient editing, where critical details remain inadequately revised, or excessive editing, which distorts the original intent or introduces unnecessary changes, which disrupts the coherence of the answer. Thus, effectively balancing the challenges of insufficient and excessive editing has become a critical issue.

\setcounter{mycounter}{2}
To address the above limitations, we propose a novel \textbf{A}tomic fact decomposition-based \textbf{R}etrieval and \textbf{E}diting (\textbf{ARE}) framework for AQA (As shown is Figure \ref{fig:introduction} (\Roman{mycounter})). Specifically, ARE first prompts LLMs to generate long-form answers for the given question. Subsequently, ARE utilizes an instruction-tuned LLM\footnote{The fine-tuned atomic fact decomposition LLM can be found in https://github.com/zhichaoyan11/FIDES}, trained on a well-constructed fact decomposition dataset, to decompose the long-form answers into molecular clauses and then into atomic facts. These atomic facts are then used to search the evidence from a search engine.

Additionally, ARE employs an evidence verifier to classify the relationships between the evidence and the atomic facts into three categories: 1) supportive, 2) editing required, and 3) irrelevant. When an atomic fact requires editing or is irrelevant to the retrieved evidence, ARE uses LLMs to process the atomic fact accordingly: if the relationship is \textit{editing required},  ARE revises the fact; if the relationship is \textit{irrelevant}, ARE expands the atomic fact for re-retrieval and verification. This process is repeated until the verification result shows that the relationship  is ``\textit{supportive}'' or the maximum number of iterations is reached. 

Finally, ARE backtracks the edited atomic facts to their original positions within the molecular clauses, forming the final revised answer while preserving the original intent. Meanwhile, the evidences are aggregated based on the relationships between the molecular clauses and atomic facts to generate the attribution report. Meanwhile, ARE introduces a more comprehensive evaluation metric $Attr_{p}$, which not only accurately assesses the precision of retrieved evidence, but also emphasizes the completeness of the evidence.
Our contributions are summarized as follows:
\begin{itemize}
    \item This paper proposes a novel \textbf{A}tomic fact decomposition-based \textbf{R}etrieval and \textbf{E}diting (\textbf{ARE}) framework for the post-hoc retrieval-based AQA task, which performs editing and verification of long-form answers at both molecular and atomic levels; 
    \item This paper introduces an innovative instruction-tuned fact decomposition LLM, which is fine-tuned on a carefully constructed molecular-to-atomic fact decomposition dataset; 
    \item This paper proposes a traceable editing method that performs atomic fact editing and evidence retrieval at the atomic level, while ensuring consistency between atomic facts and evidences, and preserving the original intent at the molecular level; 
    \item This paper designs an evaluation metric $Attr_p$ to accurately assess the precision and completeness of retrieved evidences, mitigating the proportion of invalid retrieved evidences.
\end{itemize}

\begin{figure*}[ht]

    \centering
    \includegraphics[scale=0.67]{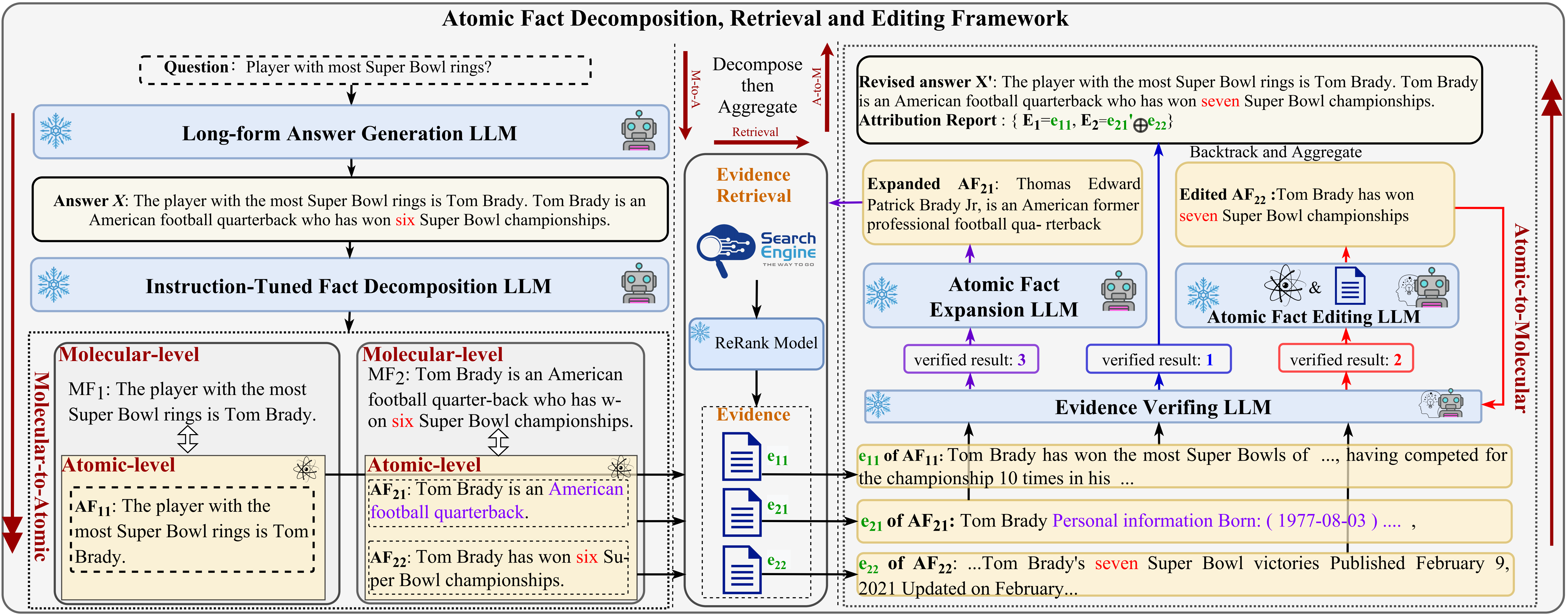}
    \caption{An example of the ARE process. In Molecular-to-Atomic stage, an instruction-tuned fact decomposition LLM decomposes the long-form answer $X$ into molecular clauses and atomic facts. Atomic facts are then used to retrieve evidences. In Atomic-to-Molecular stage, a  Verifier assesses the relationship between the evidence and facts, resulting in three states: 1. \textit{supportive}, requiring no further action; 2. \textit{editing required}, necessitating revision to the atomic facts; and 3. \textit{irrelevant}, requiring a new evidence retrieval. Finally, the atomic facts are backtracked to the original position of $X$ to generate the revised answer $X^{'}$. 
    The Attribution Report consists of all evidences $\{E_1,...,E_m\}$, where $MF_i^{'}$ and $e_{ij}^{'}$ are the edited contents.
    }
    \label{fig:framework}
\end{figure*}

\section{Related Work}
\subsection{Attributed Question Answering}
From the perspective of retrieval timing, two notable trends have recently emerged: (1) Retrieval-then-Read (RTR) and (2) Post-hoc Retrieval. Trivedi \textit{et al.} \cite{trivedi2022interleaving} introduces IRCoT, a method that interleaves retrieval with steps in a Chain of Thought (CoT). This approach not only guides the retrieval process using CoT but also utilizes the retrieved results to enhance the CoT.

Muller \textit{et al.} \cite{muller-etal-2023-evaluating} investigates attribution in cross-lingual question answering. ALCE \cite{gao2023enabling} employs various methods to integrate retrieved documents into LLMs for answer generation. Li \textit{et al.} \cite{li2023llatrieval} introduces a progressive selection of evidence using LLMs with a classification-based prompting template. Bohnet \textit{et al.} \cite{bohnet2022attributed} conducts an extensive evaluation of LLM attributions, finding that while the Retrieval-then-Read (RTR) approach performs well, it necessitates the comprehensive use of a traditional training set, thereby highlighting the potential of post-hoc retrieval. RARR \cite{gao2023rarr} is the first framework to implement question decomposition-based retrieval followed by revision. Building on RARR, PURR presents an end-to-end editor for text revision \cite{chen2023purr}. Kang \textit{et al.} \cite{kang2023mitigating} proposes a combination of RTR and post-hoc retrieval strategies. The limitations of question decomposition were discussed in the Introduction. Our approach addresses these limitations through molecular-to-atomic fact decomposition and atomic-to-molecular editing, resulting in significant performance improvements.

\subsection{Question or Fact Decomposition}
 The technology of decomposition has been shown to effectively address complex questions, particularly in various Question Answering (QA) tasks (e.g., Multi Hop Question Answering (MHQA))  \cite{lin2023decomposing, Ye_2023, zhang2023reasoning,xu-etal-2025-log} and claim verification tasks \cite{lin-etal-2023-decomposing, chen2023complex, kim2023kggpt, wang2023hallucination, kamoi-etal-2023-wice}. Under the MHQA task, \cite{chu-etal-2024-beamaggr, cai2024forag, zhuang-etal-2024-efficientrag, chen-etal-2024-llm,zhao-etal-2024-agr} typically decompose the original question or introduce retrieval augment generation methods to mitigate the ambiguity of questions and the model parameter knowledge limitations, respectively.  While fact decomposition can generate sub-facts that represent the original answer, it struggles to decompose long-form answers that have a complex semantic structure and aggregate evidence for sentences based on the decomposition results. This limitation becomes evident when retrieving evidence by fact to support each sentence that contains more than one fact.

\subsection{Hallucination Detection and Revision}
 
Hallucination detection \cite{chen2023hallucination, wang2023hallucination, kadavath2022language, mndler2023selfcontradictory, mishra2024fine} is a challenging yet essential task for improving the reliability of LLMs in real-world scenarios. Similarly, Zheng \textit{et al.} \cite{ZLLP2024} introduced TrustScore, the first effective evaluation metric designed to assess the trustworthiness of LLMs responses in a reference-free context. Yan \textit{et al.} \cite{yan2024corrective} proposed CRAG, which trains an evidence evaluator model to assess the quality of retrieved evidence and classify retrieval actions as ``Correct,” ``Incorrect,” or ``Ambiguous”. Different from our framework, CRAG aims to improve retrieval robustness by evaluating and refining retrieved evidence before generation and focuses on improving evidence quality rather than ensuring attribution and alignment with source material.

To tackle hallucinations, numerous model editing methods have been developed that do not involve updating the parameters of LLMs \cite{zheng-etal-2023-edit, gao2023rarr, HLTW+2023, dhuliawala2023chainofverification, wu2024updating, song2024knowledge}. In contrast to our approach, most of these methods focus exclusively on either hallucination detection or revision.

\section{Preliminary}

\textbf{Attributed Question Answering (AQA)} is a task which provides both a trustworthy answer and a reliable attribution report for a given question. Formally, given a question $q$ and a corpus of text passages $D$, the process can be defined as follows: 

\begin{equation}
    X,\ A \longleftarrow \mathcal{M}_{\text{AQA}}(q,\ D),
    \nonumber
\end{equation}
where $X$ represents the long-form answer, $A$ is an attribution report consisting of a collection of evidence, and $\mathcal{M}_{\text{AQA}}$ is a model used for the AQA task.

\noindent \textbf{Post-hoc Retrieval-based AQA} aims to enhance the reliability of LLM-generated long-form answers by further incorporating evidence retrieval, fact verification, and factual editing. A well-designed framework should maximize the attribution score while minimizing changes to the original intent.

\noindent \textbf{Symbol definitions.} The key symbols used in our framework are defined as follows: $MF_i$ denotes a molecular clause in the long-form answer $X$ that contains at least one fact subject to further decomposition; $AF_{ij}$ represents an atomic fact derived from $MF_i$ that cannot be further decomposed; $e_{ij}$ refers to the evidence retrieved for each atomic fact $AF_{ij}$; and $E_i$ denotes the aggregated evidence for the molecular clause $MF_i$, compiled from the corresponding evidence pieces $e_{ij}$.

\noindent \textbf{Entropy} measures the uncertainty or information content associated with random variables. The higher the entropy, the more uncertain or random the information is \cite{ash2012information}.
It can be calculated as:
\begin{equation}
\nonumber
H(X) = -\sum_{i = 1}^{n}P(x_i)\log P(x_i),
\end{equation}
where $P(x_i)$ is the probability of $x_i$, which can be defined as: 
\begin{equation}
    P(x_i) = P(t_1) \cdot P(t_2 \mid t_1) \cdot \dots \cdot P(t_i \mid t_1, t_2, \dots, t_{i-1}),
    \nonumber
\end{equation}
where $t_i$ is $i$-th token in $x_i$.

\setcounter{mycounter}{1}
\noindent \textbf{Hypothesis. } Atomic facts exhibit lower entropy than sub-questions, leading to improved evidence retrieval. 
As demonstrated by Passalis \textit{et al.} \cite{7439840} that minimizing entropy can enhance performance in retrieval tasks.
Long-form answers, characterized by greater complexity, have higher entropy, which makes it more challenging to retrieve relevant evidences. Traditional post-hoc retrieval based AQA methods use LLMs to generate sub-questions directly, which may contain multiple meanings, uncertainties and lose some key entities. Thus, these sub-questions often have high entropy, leading to retrieving ambiguous or irrelevant evidence (As shown in Figure \ref{fig:introduction} (\Roman{mycounter})). In contrast, fact decomposition preserves higher certainty and lowers entropy, thereby enhancing evidence retrieval \cite{ash2012information}.

To support this hypothesis, we begin with the following assumptions:

1) \(X\) is the long-form answer, \(\mathrm{h}\) is a deterministic extractor that maps each \(X\) to a set of entities: \(E=\mathrm{h}(X)\), hence by the definition of conditional entropy for a deterministic function, \(H(E|X)=0\).

2) The support set for atomic fact \(AF\) is \(S_{AF}(X)=\{af:P(AF=af|X)>0\}\) and for sub-question \(S_Q(X)={q:P(Q=q|X)>0}\). To fully cover the same triple fact \((e_h,r,e_t)\), the LLMs must generate at least \textit{two} sub-questions (one for head entity and one for tail). 
Hence 
\begin{equation}
    \mid S_Q(X) \mid \geq 2 \mid S_{AF}(X) \mid
    \label{eq1}
\end{equation}
3) From atomic fact set \(AF\), \(E\) can be recovered, the entropy \(H(E|AF) =0 \), similarly, from sub-question set \(Q\), the entropy \(H(E|Q)=0\).

\textbf{The theory of maximum entropy}: For a length of support set \(\mid \mathrm{supp(P)\mid}=M \), the upper bound of entropy is \(H(P) \leq \mathrm{log} M\). The generation process of the support set is considered without any bias, the maximum entropy is uniform distribution. Therefore, we adopt \(H(AF|X)\approx \mathrm{log} \mid S_{AF}(X) \mid\) and \(H(Q|X) \approx \mathrm{log} \mid S_Q(X) \mid\).

From eq \ref{eq1}, 
\begin{equation}
    \mathrm{log} \mid S_{Q} \mid \geq \mathrm{log}(2 \mid S_{AF}\mid) = \mathrm{log}2 + \mathrm{log} \mid S_{AF} \mid,
\end{equation}
consider the upper bound of entropy, 
\begin{equation}
    H(Q|X) \geq \mathrm{log}2 + \mathrm{log} \mid S_{AF}(X) \mid \geq \mathrm{log} 2 + H(AF|X)
\end{equation}
\begin{equation}
    \begin{aligned}
        H(Q|X)-H(AF|X) \geq \mathrm{log}2 > 0 \\
    \Rightarrow H(AF|X) < H(Q|X).
    \end{aligned}
\end{equation}

\textbf{The theory of chain rule of entropy:} Given the two random variables \((A|B)\), the chain is:
\begin{equation}
    H(A,B)=H(A) + H(B|A)=H(B) + H(A|B).
    \label{eq5}
\end{equation}
\begin{equation}
\begin{aligned}
\mathrm{For} \ AF\mathrm{:} \ H(E,AF)&=H(AF)+H(E|AF) \\
    &=H(AF)+0 \\
    &=H(E)+H(AF|E) \\
    & \Rightarrow H(AF)=H(E) + H(AF|E).
    \nonumber
\end{aligned}    
\end{equation}
For \(Q\),
\begin{equation}
    \begin{aligned}
        H(E,Q)&=H(Q)+H(E|Q) \\
        & = H(Q) + 0 \\
        &=H(E)+H(Q|E) \\
        & \Rightarrow H(Q)=H(E)+H(Q|E).
        \nonumber
    \end{aligned}
\end{equation}

Due to \(H(E|X)=0\), \(H(AF)=H(E)+H(AF|X)\), \(H(Q)=H(E)+H(Q|X)\). The relation of \(H(Q)\) and \(H(AF)\) is
\begin{equation}
    \begin{aligned}
        &H(Q)-H(AF) \\
        &=[H(E)+H(Q|X)]-[H(E)+H(AF|X)] \\
        &=H(Q|X)-H(AF|X) \geq \mathrm{log}2  \Rightarrow H(Q) > H(AF).
        \nonumber
    \end{aligned}
\end{equation}
The entropy of atomic fact is lower than sub-question.
The experimental results in Section \ref{hyposis} also verify our hypothesis by demonstrating that atomic-fact based retrieval outperforms sub-question based retrieval.

\begin{figure*}[t]
    \centering
    \includegraphics[scale=1.34]{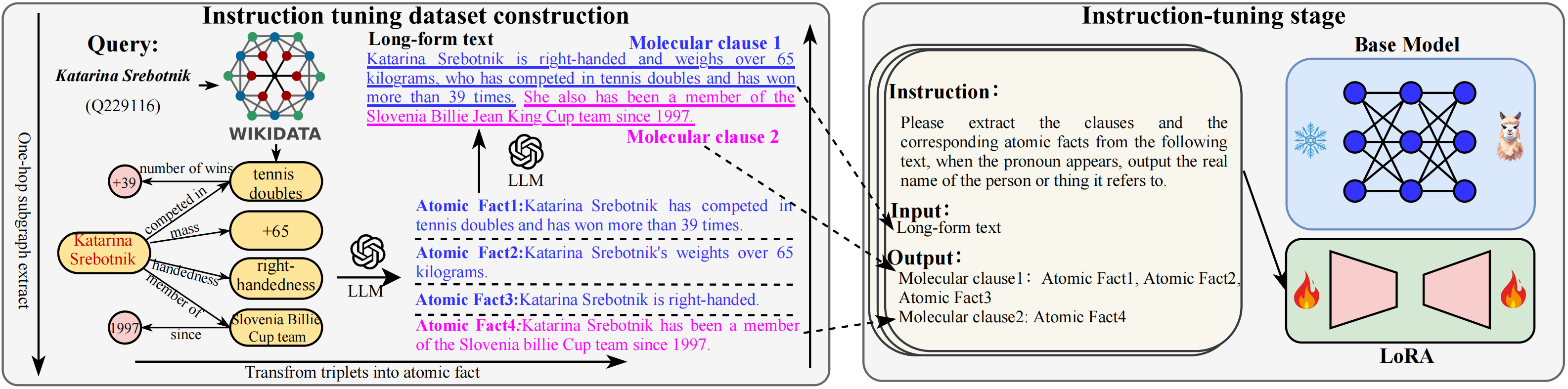}
    \caption{The process of dataset construction and instruction tuning. We first extract the neighbors of the corresponding entity from Wikidata, and then utilize an LLM to convert each  triple into a single atomic fact.  Subsequently, all facts are transformed into a text comprising multiple molecular clauses by the LLM. The atomic facts are  then used to construct each sentence. After two prompts, we obtain the dataset for instruction-tuning with LoRA.}
    \label{fig:instruct}
\end{figure*}

\section{Approach}
In this section, a novel \textbf{A}tomic fact decomposition-based \textbf{R}etrieval and \textbf{E}diting (\textbf{ARE}) framework is proposed for AQA task. Specifically, our ARE framework contains three stages: 1) \textbf{Molecular-to-Atomic Fact Decomposition} generally decomposes the generated long-form answer into molecular clauses and atomic facts by an instruction-tuned LLM; 2) \textbf{Atomic Fact based Evidence Retrieval} employs a search engine to retrieve evidence and selects the most similar evidence using a reranking model; 3) \textbf{LLM-based Evidence Verifying and Editing} guides LLMs through a verified result to either expand atomic facts for further re-retrieval or edit atomic facts with retrieved evidence.

\subsection{Molecular-to-Atomic Fact Decomposition}
\label{section1}

This stage introduces the process that prompts LLMs to generate a long-form answer, and utilizes the instruction-tuned LLMs fine-tuned  on a carefully constructed dataset to decompose the long-form answer into molecular clauses and atomic facts. 

\subsubsection{Long-form answer generation}
\label{answer generation}

LLMs are trained on the large-scale corpora which endow them with powerful generative capabilities. Long-form answers are generated by feeding a few-shot prompting template and a specific question to LLMs. Formally, the whole process can be represented as:

\begin{tcolorbox}[colback=white!5!white, colframe=black!75!black, title=\textit{Question: Player with most Super Bowl rings?}]
\textbf{Long-form Answer $X$:} \textit{The player with the most Super Bowl rings is Tom Brady. Tom Brady is an American football quarterback who has won six Super Bowl championships.}
\end{tcolorbox}

\subsubsection{Fact decomposition}
\label{Fact decomposition}
Fact decomposition aims to decompose long-form answers characterized by complex logical structures into molecular clauses and multiple atomic facts through the instruction-tuning LLM. Specifically, the fact decomposition module consists of two key components: data construction and the instruction-tuning of the Molecule-to-Atomic LLM.

\noindent \textbf{Dataset Construction.}  
To fine-tune the molecule-to-atomic fact decomposition LLM, ARE employs LLMs to construct an instruct-tuning dataset that converts triples into texts containing molecule clauses and atomic facts derived from Knowledge Graphs (KGs) \cite{wang2023survey, wang2024ime}. As depicted in Figure \ref{fig:instruct}, the dataset construction process is divided into two steps: data collection and data processing with LLMs.

\textit{Step1}: ARE first randomly selects the entities from the Knowledge Graph Question Answering (KGQA) dataset \cite{gu2021beyond} and then extracts the triplets of given entities from Wikidata\footnote{https://www.wikidata.org/w/api.php?action=wbsearchentities\&search=}. As some properties or objects in the triplets may be useless (e.g., JPEG, code, and ISSN), they are thus further filtered by heuristic rules in the extracting process.

\textit{Step2}: ARE utilizes LLMs to randomly transform several triplets into atomic facts $\{AF_{i1},...,AF_{ik}\}$. The atomic facts are fed into LLMs to generate a response in the Json format: \{``Generated content": $S$, ``Molecular clauses": $MF_i$,  ``Atomic facts": $AF_{ij}$\}, where $S$ is text consisting of $m$ molecular clauses, $MF_i$ is $i$-th molecular clause, $i \in \{1,...,m\}$, and $AF_{ij}$ is the $j$-th atomic fact of the $MF_i$, $j \in \{1,...,k\}$.To retain the overall coherence and logical flow of the original long form answer, we format the molecular clauses and atomic facts into a Json structure: \{``output'':[\{$MF_i$:$AF_{ij}...$\}...]\}, which enhances the relationship among $MF_i$, $AF_{ij}$ and $X$.

Through the above process, an instruction-tuning dataset can be obtained, which contains 7138 samples for training and 1000 for evaluation. 

\noindent \textbf{Instruction-tuning the Molecule-to-Atomic LLM.} For the fine-tuning stage, ARE employs the Low-Rank Adaptation (LoRA) technique \cite{hu2022lora} for instruction-tuning the Llama3-8B-Instruct LLM to reduce computational complexity and time consumption.  

Specifically, LoRA works by freezing parameters $\theta_0$ of the pre-trained model, while adding trainable parameters $\Delta \theta_0$ that can be expressed as the product of two low-rank matrices:
\begin{equation}
    \Delta \theta_0 = \mathbf{BA},
\end{equation}
where $\mathbf{B}\in \mathbb{R}^{d\times r} $, $\mathbf{A} \in \mathbb{R}^{r\times k}$, $r\ll \mathrm{min}(d,k)$.

After obtaining the instruction-tuned Molecule-to-Atomic LLM, ARE decomposes the long-form answer ``$X$: \textit{The player with the most Super Bowl rings is Tom Brady. Tom Brady is an American football quarterback who has won six Super Bowl championships.}'' into molecular clauses ``$MF_1$: \textit{The player with the most Super Bowl rings is Tom Brady.}'' and ``$MF_2$: \textit{Tom Brady is an American football quarterback who has won six Super Bowl championships.}'', $MF_1$ only has one atomic fact $AF_{11}$, $MF_2$ can be further decomposed into two atomic facts ``$AF_{21}$: \textit{Tom Brady is an American football quarterback.}'' and ``$AF_{22}$: \textit{Tom Brady has won six Super Bowl championships.}''.

\subsection{Atomic Fact based Evidence Retrieval}
\label{section2}

Atomic fact based evidence retrieval aims to employ search engines\footnote{We select Google Search as knowledge source, accessible via https://customsearch.googleapis.com/customsearch/v1} to search evidences that support atomic facts. 
To be specific, ARE utilizes a pre-trained Sentence-Bert model to generate embeddings for atomic facts and their corresponding evidences, and subsequently re-ranks evidences based on similarity to identify the most relevant one.

ARE leverages search engines to search evidence from each atomic fact $AF_{ij}$, while uses the pre-trained Sentence-Bert model\footnote{https://huggingface.co/cross-encoder/ms-marco-MiniLM-L-6-v2} to embed both the searched evidences $e_c$ and the atomic fact $AF_{ij}$ into a common vector space:

\begin{equation}
    \mathbf{e}_c^z, \text{\textbf{AF}}_{ij} = \textrm{Sentence-Bert}(e_c^z, AF_{ij}),
\end{equation}
where $e_c^z, z \in \{1,...,o\}$ is the $z$-th searched evidence in $e_c$. ARE calculates the relevance score $R$($\mathbf{e}_c^z$, $\text{\textbf{AF}}_{ij}$) through the cosine similarity function:
\begin{equation}
    R(\mathbf{e}_c^z, \text{\textbf{AF}}_{ij}) = \frac{\mathbf{e}_c^z\cdot \text{\textbf{AF}}_{ij}}{\parallel \mathbf{e}_c^z\parallel\parallel\text{\textbf{AF}}_{ij}\parallel},
\end{equation}
where ``$\cdot$'' represents the inner product and $\parallel \cdot \parallel$ is the norm of the corresponding vector \cite{wang2024large}. Finally, ARE ranks all the evidences $\{e_c^1,...,e_c^o\}$ of $AF_{ij}$ by relevance score $R$($\mathbf{e}_c^z$, $\text{\textbf{AF}}_{ij}$) and selects the top as the most relevant evidence $e_{ij}$ for $AF_{ij}$. 
The search engine's response may not always yield valid content, often due to poorly constructed queries or limitations within the search engine itself.

\subsection{LLM-based Evidence Verifying and Editing}
\label{section3}

Given that hallucinated content may exist in long-form answers generated by LLMs, ARE employs an evidence-verifying LLM to assess the factual consistency of atomic facts. This verifier operates using a few-shot, prompt-based approach to determine whether an atomic fact should be retained, edited, or re-retrieved, based on a comparison with the corresponding retrieved evidence. The verifier takes as input an atomic fact along with its associated evidence and outputs one of several status labels, including:
\begin{equation}
    \textrm{EV}(AF_{ij},e_{ij})=\left\{\begin{matrix}
  1, & \textit{supportive} \\ 
  2, & \textit{editing required}\\
  3, & \textit{irrelevant}  ,   
\end{matrix}\right.
\end{equation}
where $\text{EV}(\cdot,\cdot)$ denotes the evidence verifying LLM, which determines the relationships between the atomic fact and evidence. Specifically, $1$ represents \textit{supportive}, requiring no further action; $2$ means \textit{editing required}, necessitating the revision of atomic facts; $3$ indicates \textit{irrelevant}, which requires a new evidence retrieval. To address potential misjudgments arising from LLM uncertainty, we implement an iterative verification mechanism. This process repeatedly validates results flagged with the statuses \textit{editing required} and \textit{irrelevant}, ensuring greater accuracy in the edited or re-retrieved content. The prompt template used by the evidence verifier is provided in Table \ref{tab:prompt}.

\begin{table}[t]
\centering
\label{tab:}
\resizebox{0.485\textwidth}{!}{%
\begin{tabular}{rl}
\hline
\multicolumn{2}{l}{\begin{tabular}[c]{@{}l@{}}Algorithm 1: Atomic fact decomposition-based  Retrieval and \\ Editing framework.\end{tabular}} \\ \hline
\multicolumn{1}{r}{\begin{tabular}[c]{@{}r@{}}1:\\ 2:\\ 3:\\ 4:\\ 5:\\ 6:\\ 7:\\ 8:\\ 9:\\ 10:\\ 11:\\ 12:\\ 13:\\ 14:\\ 15:\\ 16:\\ 17:\\ 18:\\ 19:\\ 20:\end{tabular}} & \begin{tabular}[l]{@{}l@{}}
\textbf{Input:} The question \textit{q}, the corpus of text passages \textit{D}; \\  
$A \leftarrow $ \{\}; \\  
$X$ $\leftarrow$ LLMs(q); \\  
$MF$, $AF$ $\leftarrow$ Fact Decomposition($X$); \\  
\textbf{for} $MF_i \in MF$ \textbf{do} \\  
\ \ \textbf{for} $AF_{ij} \in AF$ \textbf{do} \\  
\ \ \ \ $e_{ij}$ $\leftarrow$ Evidence Retrieval($AF_{ij}$, $D$); \\  
\ \ \ \ \textbf{for} iteration=0: 4 \textbf{do} \\  
\ \ \ \ \ \ \ \ status $\leftarrow$ Evidence Verifying($AF_{ij}$, $e_{ij}$); \\  
\ \ \ \ \ \ \ \ \textbf{if} status == 1 \textbf{do} \\ 
\ \ \ \ \ \ \ \ \ \ $AF[j] \leftarrow AF_{ij}$ \\  
\ \ \ \ \ \ \ \ \ \ break \\  
\ \ \ \ \ \ \ \ \textbf{if} status == 2 \textbf{do} \\  
\ \ \ \ \ \ \ \ \ \ $AF_{ij}$ $\leftarrow$ Editing($AF_{ij}$, $e_{ij}$) \\ 
\ \ \ \ \ \ \ \ \textbf{if} status == 3 \textbf{do} \\  
\ \ \ \ \ \ \ \ \ \ \ Expanded $AF_{ij}$ $\leftarrow$ Fact Expansion($AF_{ij}$) \\  
\ \ \ \ \ \ \ \ \ \ \ $e_{ij} \leftarrow$ Evidence Retrieval(Expanded $AF_{ij}$, $D$); \\  
\ \ $A\  \cup$ Aggregate($e_{ij}$) \\  
$X' \leftarrow$ Backtrack($AF$, $MF$) \\  
\textbf{output}: The revised answer $X'$ and the attribution report $A$.\end{tabular} \\ \hline
\end{tabular}%
}
\end{table}

Figure \ref{fig:framework} illustrates the process. The status between $AF_{11}$ and $e_{11}$ is classified as \textit{supportive}, $AF_{21}$ retrieves irrelevant evidence $e_{21}$, thus requiring re-retrieval. $AF_{22}$ needs to be edited based on evidence $e_{22}$. For \textit{irrelevant}, since LLMs can provide relevant information to guide retrieval systems \cite{wang-etal-2023-query2doc}, ARE utilizes the fact expansion LLM to expand the atomic fact $AF_{21}$. The expanded fact contains more complete contextual information, enabling a more effective retrieval of relevant evidence. For example, ``\textit{Tom Brady}'' is expanded to ``\textit{Thomas Edward Patrick Brady Jr.}'' which is a more complete entity. The newly retrieved evidence $e_{ij}'$ will be re-verified by the verifier, and this process will be repeated. The details of the prompt template can be found in Table \ref{tab:prompt}.

For \textit{editing required}, ARE designs prompts for LLMs based on in-context learning and chain-of-thought prompting techniques ~\cite{zheng-etal-2023-edit,gao2023rarr}, guiding LLMs to revise the atomic fact $AF_{22}$ using the retrieved evidence $e_{22}$. 
Since the edits are made at the atomic fact level rather than revising the entire answer $X$, ARE can precisely adjust the necessary details, thus minimizing unnecessary modification. Detailed editing instructions are shown in Table \ref{tab:prompt}.

The edited atomic facts are re-verified and reintegrated into their original positions within the molecular clauses, producing the final revised long-form answer $X'$. This process also preserves molecular clauses that did not require editing. During backtracking, both molecular and atomic facts are provided in Json format alongside the original long-form answer as a reference. This approach follows the instruction-tuning dataset construction methodology, enabling the LLM to effectively manage interdependencies among atomic facts. The detailed backtracking prompt is presented in Table \ref{tab:prompt}. 

To further obtain the attribution report $A$, all evidences $e_{ij}$ are aggregated into a sequence $E_i$ to support $MF_i$. Since each $AF_{ij}$ is derived from the decomposition of $MF_i$, there may be overlaps among the evidence $e_{ij}$. To address this issue, duplicate snippets are removed. Ultimately, $A$ is compiled as $\{E_1,...,E_i,...,E_m \}$. 
To intuitively demonstrate the entire procedure of our proposed method, we have summarized the process in Algorithm 1.

\begin{table*}[ht]
\centering
\caption{Evaluation results on API-based GPT-3.5 and GPT-4o-mini. When evaluating different LLMs, all methods also use the corresponding LLM for construction. \textit{$Attr_r$} and $Attr_p$ are used to evaluate the attribution, while \textit{Pres} is used to assess the Preservation after editing. $F1_{PP}$ and $F1_{RP}$ are the harmonic means of the \textit{$Attr_p$} and \textit{Pres}, and the \textit{$Attr_r$} and \textit{Pres} metrics, respectively. Additionally, Average $\mathrm{F1_{PP}}$ (Ave-$\mathrm{F1_{PP}}$) and $\mathrm{F1_{RP}}$ (Ave-$\mathrm{F1_{RP}}$) are used to provide a comprehensive overview for each row. Note that the RARR paper does not provide access to the test datasets; thus, the reported results are reproduced from the publication. Bold indicates the best performance, while underlining indicates the second-best performance.}
\label{tab:main1}
\setlength{\tabcolsep}{0.28cm}{
\begin{tabular}{cccc:cc|ccc:cc|cc}
\hline
 \multirow{2}{*}{\textbf{Methods}}& \multicolumn{5}{c}{\textbf{GPT-3.5}} & \multicolumn{5}{c}{\textbf{GPT-4o-mini}} & \textbf{} & \textbf{} \\
 \cline{2-13}
 & $Attr_r$ & $Attr_p$ & Pres & $\mathrm{F1_{PP}}$ & $\mathrm{F1_{RP}}$ & $Attr_r$ & $Attr_p$ & Pres & $\mathrm{F1_{PP}}$ & $\mathrm{F1_{RP}}$ & Ave-$\mathrm{F1_{PP}}$ & Ave-$\mathrm{F1_{RP}}$ \\ \hline 
{} & \multicolumn{12}{c}{\cellcolor[HTML]{BFBFBF}\textbf{NQ}} \\
DRQA & 0.424 & 0.647 & - & - & - & 0.265 & 0.576 & - & - & - & - & - \\
EFEC & 0.598 & 0.042 & 0.762 & 0.080 & 0.670 & 0.492 & 0.053 & 0.688 & 0.098 & 0.546 & 0.089 & 0.608 \\
CCVER & 0.624 & 0.066 & 0.928 & \underline{0.123} & \underline{0.747} & 0.563 & 0.051 & 0.931 & 0.097 & \underline{0.702} & 0.110 & \underline{0.728} \\ 
RARR & 0.649 & 0.058 & 0.868 & 0.109 & 0.743 & 0.493 & 0.080 & 0.904 & \underline{0.147} & 0.638 & \underline{0.128} & 0.691 \\
\textbf{ARE} & 0.670 & 0.756 & 0.910 & \textbf{0.826} & \textbf{0.772} & 0.623 & 0.722 & 0.902 & \textbf{0.802} & \textbf{0.737} & \textbf{0.814} & \textbf{0.755} \\ \cline{2-13} 
 & \multicolumn{12}{c}{\cellcolor[HTML]{BFBFBF}\textbf{Mintaka}} \\
DRQA & 0.431 & 0.673 & - & - & - & 0.251 & 0.544 & - & - & - & - & - \\
EFEC & 0.557 & 0.040 & 0.729 & 0.076 & 0.632 & 0.497 & 0.049 & 0.698 & 0.092 & 0.581 & 0.084 & 0.665 \\
CCVER & 0.630 & 0.069 & 0.937 & \underline{0.129} & \underline{0.753} & 0.634 & 0.045 & 0.933 & 0.086 & \textbf{0.755} & 0.108 & \underline{0.754} \\
RARR & 0.646 & 0.060 & 0.850 & 0.112 & 0.734 & 0.508 & 0.058 & 0.899 & \underline{0.109} & 0.649 & \underline{0.111} & 0.692 \\
\textbf{ARE} & 0.716 & 0.807 & 0.914 & \textbf{0.857} & \textbf{0.803} & 0.634 & 0.755 & 0.881 & \textbf{0.813} & \underline{0.737} & \textbf{0.835} & \textbf{0.770} \\ \cline{2-13} 
 & \multicolumn{12}{c}{\cellcolor[HTML]{BFBFBF}\textbf{StrategyQA}} \\
DRQA & 0.237 & 0.490 & - & - & - & 0.080 & 0.222 & - & - & - & - & - \\
EFEC & 0.354 & 0.049 & 0.716 & 0.092 & 0.474 & 0.330 & 0.071 & 0.638 & 0.128 & 0.435 & 0.110 & 0.455 \\
CCVER & 0.372 & 0.047 & 0.932 & 0.089 & \underline{0.532} & 0.341 & 0.056 & 0.961 & 0.106 & \underline{0.503} & 0.098 & \underline{0.518} \\
RARR & 0.356 & 0.097 & 0.862 & \underline{0.174} & 0.504 & 0.292 & 0.079 & 0.916 & \underline{0.145} & 0.443 & \underline{0.160} & 0.474 \\
\textbf{ARE} & 0.463 & 0.559 & 0.899 & \textbf{0.689} & \textbf{0.611} & 0.413 & 0.469 & 0.904 & \textbf{0.618} & \textbf{0.567} & \textbf{0.654} & \textbf{0.589} \\ \cline{2-13} 
 & \multicolumn{12}{c}{\cellcolor[HTML]{BFBFBF}\textbf{ExpertQA}} \\
DRQA & 0.127 & 0.283 & - & - & - & 0.122 & 0.308 & - & - & - & - & - \\
EFEC & 0.343 & 0.071 & 0.686 & 0.129 & 0.457 & 0.297 & 0.075 & 0.635 & 0.134 & 0.405 & 0.132 & 0.431 \\
CCVER & 0.292 & 0.071 & 0.967 & 0.132 & 0.449 & 0.310 & 0.062 & 0.980 & 0.117 & \underline{0.471} & \underline{0.221} & \underline{0.460} \\
RARR & 0.340 & 0.078 & 0.904 & \underline{0.144} & \underline{0.494} & 0.269 & 0.064 & 0.944 & \underline{0.120} & 0.419 & 0.132 & 0.457 \\
\textbf{ARE} & 0.412 & 0.438 & 0.917 & \textbf{0.593} & \textbf{0.569} & 0.387 & 0.414 & 0.914 & \textbf{0.570} & \textbf{0.544} & \textbf{0.582} & \textbf{0.557} \\ \hline
\end{tabular}%
}
\end{table*}

\begin{table*}[ht]
\centering
\caption{Evaluation results on the open-source Llama3-70B and Llama2-70B. The metrics and other experimental settings are the same as those used in API-based evaluation. Bold indicates the best performance, while underlining indicates the second-best performance.}
\label{tab:main2}
\setlength{\tabcolsep}{0.28cm}{
\begin{tabular}{cccc:cc|ccc:cc|cc}
\toprule
 \multirow{2}{*}{\textbf{Methods}} & \multicolumn{5}{c}{\textbf{Llama3-70B}} & \multicolumn{5}{c}{\textbf{Llama2-70B}} & \textbf{} & \textbf{} \\
 \cline{2-13}
 & $Attr_r$ & $Attr_p$ & Pres & $\mathrm{F1_{PP}}$ & $\mathrm{F1_{RP}}$ & $Attr_r$ & $Attr_p$ & Pres & $\mathrm{F1_{PP}}$ & $\mathrm{F1_{RP}}$ & Ave-$\mathrm{F1_{PP}}$ & Ave-$\mathrm{F1_{RP}}$ \\ \hline 
{} & \multicolumn{12}{c}{\cellcolor[HTML]{BFBFBF}\textbf{NQ}} \\
DRQA & 0.382 & 0.620 & - & - & - & 0.483 & 0.522 & - & - & - & - & - \\
EFEC & 0.490 & 0.058 & 0.717 & 0.107 & 0.582 & 0.357 & 0.058 & 0.719 & 0.107 & 0.477 & 0.107 & 0.530 \\
CCVER & 0.597 & 0.071 & 0.921 & \underline{0.132} & 0.724 & 0.423 & 0.068 & 0.813 & \underline{0.126} & 0.539 & \underline{0.129} & 0.632 \\
RARR & 0.646 & 0.060 & 0.850 & 0.112 & \underline{0.734} & 0.516 & 0.066 & 0.594 & 0.119 & \underline{0.552} & 0.115 & \underline{0.643} \\
\textbf{ARE} & 0.682 & 0.739 & 0.898 & \textbf{0.811} & \textbf{0.759} & 0.584 & 0.682 & 0.926 & \textbf{0.785} & \textbf{0.716} & \textbf{0.798} & \textbf{0.738} \\ \cline{2-13} 
 & \multicolumn{12}{c}{\cellcolor[HTML]{BFBFBF}\textbf{Mintaka}} \\
DRQA & 0.380 & 0.640 & - & - & - & 0.368 & 0.600 & - & - & - & - & - \\
EFEC & 0.538 & 0.057 & 0.728 & 0.106 & 0.619 & 0.498 & 0.029 & 0.739 & 0.056 & 0.595 & 0.081 & 0.607 \\
CCVER & 0.582 & 0.073 & 0.901 & \underline{0.135} & 0.707 & 0.397 & 0.036 & 0.850 & 0.069 & 0.541 & 0.102 & 0.624 \\
RARR & 0.651 & 0.065 & 0.829 & 0.121 & \underline{0.729} & 0.543 & 0.058 & 0.679 & \underline{0.107} & \underline{0.603} & \underline{0.114} & \underline{0.666} \\
\textbf{ARE} & 0.712 & 0.767 & 0.887 & \textbf{0.823} & \textbf{0.790} & 0.631 & 0.706 & 0.940 & \textbf{0.806} & \textbf{0.755} & \textbf{0.815} & \textbf{0.773} \\ \cline{2-13} 
 & \multicolumn{12}{c}{\cellcolor[HTML]{BFBFBF}\textbf{StrategyQA}} \\
DRQA & 0.237 & 0.379 & - & - & - & 0.234 & 0.467 & - & - & - & - & - \\
EFEC & 0.319 & 0.051 & 0.666 & 0.095 & 0.432 & 0.361 & 0.031 & 0.721 & 0.059 & 0.481 & 0.077 & 0.457 \\
CCVER & 0.435 & 0.063 & 0.917 & 0.118 & \underline{0.590} & 0.323 & 0.058 & 0.854 & \underline{0.109} & 0.468 & 0.113 & 0.529 \\
RARR & 0.449 & 0.073 & 0.846 & \underline{0.134} & 0.586 & 0.412 & 0.057 & 0.604 & 0.104 & \underline{0.490} & \underline{0.119} &  \underline{0.538} \\
\textbf{ARE} & 0.474 & 0.502 & 0.907 & \textbf{0.646} & \textbf{0.623} & 0.484 & 0.533 & 0.912 & \textbf{0.673} & \textbf{0.633} & \textbf{0.660} & \textbf{0.628} \\ \cline{2-13} 
 & \multicolumn{12}{c}{\cellcolor[HTML]{BFBFBF}\textbf{ExpertQA}} \\
DRQA & 0.131 & 0.326 & - & - & - & 0.147 & 0.319 & - & - & - & - & - \\
EFEC & 0.356 & 0.076 & 0.698 & 0.137 & 0.472 & 0.357 & 0.058 & 0.719 & 0.107 & 0.477 & 0.122 & 0.475 \\
CCVER & 0.282 & 0.081 & 0.942 & 0.149 & 0.434 & 0.212 & 0.064 & 0.845 & 0.119 & 0.339 & 0.134 & 0.387 \\
RARR & 0.400 & 0.084 & 0.851 & \underline{0.153} & \underline{0.544} & 0.353 & 0.074 & 0.606 & \underline{0.132} & \underline{0.446} & \underline{0.142} & \underline{0.495} \\
\textbf{ARE} & 0.425 & 0.417 & 0.924 & \textbf{0.575} & \textbf{0.582} & 0.390 & 0.386 & 0.942 & \textbf{0.548} & \textbf{0.552} & \textbf{0.561} & \textbf{0.567} \\ \bottomrule
\end{tabular}%
}
\end{table*}

\section{Experiments}

\subsection{Evaluation Setups}

\noindent \textbf{Dataset}. We perform extensive experiments on three widely-used Question Answering (QA) datasets, as well as the AQA dataset, i.e., ExpertQA.  \textbf{Natural Questions (NQ)} \cite{kwiatkowskinatural} is a dataset for answering questions developed by Google Search and is widely used to evaluate machine reading comprehension, information retrieval, and other tasks. 
 \textbf{StrategyQA} \cite{sen2022mintaka} requires models to provide the reasoning process to come up with correct answers. Because, the correct answers are typically based on different paragraphs from multiple documents. \textbf{Mintaka} \cite{geva2021strategyqa} is a complex multilingual dataset featuring question types such as superlative and multi-hop, which is aknowledge-intensive dataset and widely used to evaluate end-to-end question answering models.  \textbf{ExpertQA} \cite{malaviya-etal-2024-expertqa} encompasses 7 question types across 32 domains and is specifically designed to evaluate attribution-based question answering models.

These datasets are used to evaluate the attribution ability in knowledge-intensive QA task. Following the standard dataset settings \cite{gao2023rarr}, we randomly select 150 samples from NQ, Mintaka, and StrategyQA as test datasets. For ExpertQA, we use the entire provided test set.

\noindent \textbf{Baselines}. The proposed ARE is compared with four post-hoc retrieval based baselines, including: 
\textbf{DRQA} \cite{bohnet2022attributed} is a straightforward attribution model that concatenates the question and answer into a single query before retrieval.\textbf{EFEC}\footnote{https://github.com/j6mes/acl2021-factual-error-correction} \cite{thorne-vlachos-2021-evidence} is an editor built on the T5 model and modifies a claim by incorporating evidence.
\textbf{RARR}\footnote{https://github.com/anthonywchen/RARR} \cite{gao2023rarr} first generates a series of sub-questions based on the long-form answer and retrieve evidence using these sub-questions as queries. The evidence will be used to verify the claim and further revise it by LLMs. \textbf{CCVER}\footnote{https://github.com/jifan-chen/Fact-checking-via-Raw-Evidence} \cite{chen-etal-2024-complex} is a model that focusses on fact verification, it uses a few-shot approach to prompt LLMs to generate ``Yes or No'' type sub-questions for a given claim, which are then used as queries to retrieve evidence.

To ensure a fair comparison across different baselines, we use the same retrieval settings for EFEC as we did for RARR, and we employ RARR's editing module for CCVER, retaining the sub-questions generation prompt of CCVER to examine the impact of generating various types of questions on attribution performance.

\noindent \textbf{LLMs.} We evaluated four LLMs: two open-source models (Llama2-70B-Chat, Llama3-70B-Instruct) and two API-based models (GPT-3.5-Turbo-1106, GPT-4o-mini-2024-07-18). For each specific task (e.g., long-form answer generation), the same model version was used across experiments, with variations only in prompt templates tailored to the task. The only exception is our instruction-tuned LLama3-8B-Instruct model, which is specifically fine-tuned for the fact decomposition task.

\noindent \textbf{Metrics.} We assess the attribution and editing through several metrics: $Attr_r$ is a molecular-level attribution metric, which measures the recall of retrieved evidences \cite{gao2023rarr}:
\begin{equation}
  Attr_r(X, A)= \underset{MF_i \in X}{\textrm{avg}} \  \underset{E_i \in A}{\textrm{max}}\ \mathrm{NLI}(E_i,MF_i), \label{attr}
\end{equation}
where, $MF_i \in X $ is a molecular clause and $E_i \in A$ is the evidence for $MF_i$, ``max'' selects the highest entailment score among all evidences, and ``avg'' calculates the average score of all evidence, $\mathrm{NLI}(E_i,MF_i)$\footnote{https://huggingface.co/google/t5\_xxl\_true\_nli\_mixture} represents the model probability of \textit{$E_i$} entailing \textit{$MF_i$}. The NLI model is based on T5-11B, which is trained on six datasets: MNLI, SNLI, FEVER, PAWS, SciTail, and VitaminC ~\cite{williams-etal-2018-broad,bowman-etal-2015-large,thorne-etal-2018-fever,zhang-etal-2019-paws,khot2018scitail,schuster-etal-2021-get}. The input format is: ``premise: \{evidence\} hypothesis: \{sentences\}''. As suggested by \cite{gao2023rarr}, we use the probability whose predicting result is ``1'' as the entailment score in the metric $Attr_r$.

However, $Attr_r$ focuses solely on evidence recall, neglecting the precision of the recalled evidence. This can result in high scores when the quantity of evidence is sufficiently large. To this end, we propose $Attr_{p}$ to simultaneously assess the precision and completeness of the evidences by calculating the proportion of invalid evidence among all evidences: 
\begin{equation}
\begin{aligned}
     Attr_{p} = \frac{\sum_{i = 1}^{m}\mathbb{I}({\textrm{NLI}_{bi}(E_i,MF_i)})}{m}, \label{precision}
\end{aligned}
\end{equation}
where $\mathbb{I}(\mathrm{condition})$ is 1 if the condition is true, 0 otherwise; $E_i \in A$ and $MF_i \in X$. The aggregated evidence \(E_i\) for a molecular clause $MF_i$ consists of multiple evidence snippets $e_{ij}$, each corresponding to an atomic fact $AF_{ij}$. We define \(E_i\) as ``invalid'' if any individual snippet $e_{ij}$ fails to support its associated atomic fact. 

To assess this, we use $\mathrm{NLI}_{bi}(\cdot,\cdot)$, a binary classification model based on the TRUE framework. Unlike the probabilistic version $\mathrm{NLI}(\cdot,\cdot)$, $\mathrm{NLI}_{bi}(\cdot,\cdot)$ returns ``1'' only if the evidence fully supports the entire sentence. This makes $Attr_p$ a stricter metric than $Attr_r$, enabling it to better evaluate the completeness of the retrieved evidence.

Preservation \cite{gao2023rarr} generally utilizes Levenshtein distance to measure the changed information from $X$ to $X'$:
\begin{equation}
  Pres_{(X,X')}=\textrm{max}(1-\frac{\textrm{Lev}(X,X')}{\textrm{length}(X)},0),
\end{equation}
where $Pres_{(X,X')}$ equals 1 when
$X$ is identical to $X'$, indicating no changes. A value of 0 means $X$ and $X'$ do not share common words, reflecting a complete divergence.

To better compare with different baselines, $F1_{RP}$ \cite{gao2023rarr,chen2023purr} and $F1_{PP}$ are proposed. $F1_{RP}$ and $F1_{PP}$ are calculated by the following equations:
\begin{equation}
    \mathrm{F1_{RP} = \frac{2*Attr_r*Pres}{Attr_r+Pres}},\quad
    \mathrm{F1_{PP} = \frac{2*Attr_p*Pres}{Attr_p+Pres}}.
\end{equation}

\begin{table*}[h]
\centering
\caption{The ablation experiment results on NQ, StrategyQA, Mintaka and ExpertQA using GPT-3.5. ``\textit{w/o}'' represents removal for the mentioned module. We mark the better results in bolded.}
\label{tab:ablation}
\setlength{\tabcolsep}{0.4cm}{
\begin{tabular}{cccc:cc|ccc:cc}
\toprule
\multirow{2}{*}{\textbf{Method}} & \multicolumn{5}{c}{\cellcolor[HTML]{BFBFBF}\textbf{NQ}} & \multicolumn{5}{c}{\cellcolor[HTML]{BFBFBF}\textbf{StragegyQA}} \\ \cline{2-11} 
 & $Attr_r$ & $Attr_p$ & Pres & $\mathrm{F1_{PP}}$ & $\mathrm{F1_{RP}}$ & $Attr_r$ & $Attr_p$ & Pres & $\mathrm{F1_{PP}}$ & $\mathrm{F1_{RP}}$ \\ \hline
\textit{w/o} edit & 0.648 & 0.715 & - & - & - & 0.442 & 0.53 & - & - & - \\
\textit{w/o} atomic level & 0.451 & 0.619 & 0.9 & 0.734 & 0.608 & 0.225 & 0.331 & 0.918 & 0.487 & 0.362 \\
\textit{w/o} molecular level & 0.713 & 0.785 & 0.585 & 0.670 & 0.643 & 0.514 & 0.609 & 0.598 & 0.603 & 0.553 \\
\textit{w/o} re-retrieval & 0.609 & 0.677 & 0.911 & 0.777 & 0.730 & 0.437 & 0.533 & 0.927 & 0.677 & 0.594 \\
\textbf{ARE} & 0.69 & 0.737 & 0.91 & \textbf{0.814} & \textbf{0.772} & 0.463 & 0.559 & 0.899 & \textbf{0.689} & \textbf{0.611} \\ \hline
 & \multicolumn{5}{c}{\cellcolor[HTML]{BFBFBF}\textbf{Mintaka}} & \multicolumn{5}{c}{\cellcolor[HTML]{BFBFBF}\textbf{ExpertQA}} \\ \cline{2-11} 
\textit{w/o} edit & 0.672 & 0.768 & - & - & - & 0.404 & 0.417 & - & - & - \\
\textit{w/o} atomic level & 0.477 & 0.651 & 0.928 & 0.765 & 0.630 & 0.168 & 0.308 & 0.943 & 0.464 & 0.285 \\
\textit{w/o} molecular level & 0.725 & 0.832 & 0.548 & 0.661 & 0.624 & 0.425 & 0.449 & 0.561 & 0.499 & 0.484 \\
\textit{w/o} re-retrieval & 0.68 & 0.767 & 0.936 & 0.843 & 0.788 & 0.388 & 0.421 & 0.937 & 0.581 & 0.548 \\
\textbf{ARE} & 0.716 & 0.807 & 0.914 & \textbf{0.857} & \textbf{0.803} & 0.412 & 0.438 & 0.917 & \textbf{0.593} & \textbf{0.569} \\ \bottomrule
\end{tabular}%
}
\end{table*}

\begin{figure}[ht]
    \centering
    
    \begin{minipage}{\linewidth}
        \centering
        \includegraphics[scale=0.2]{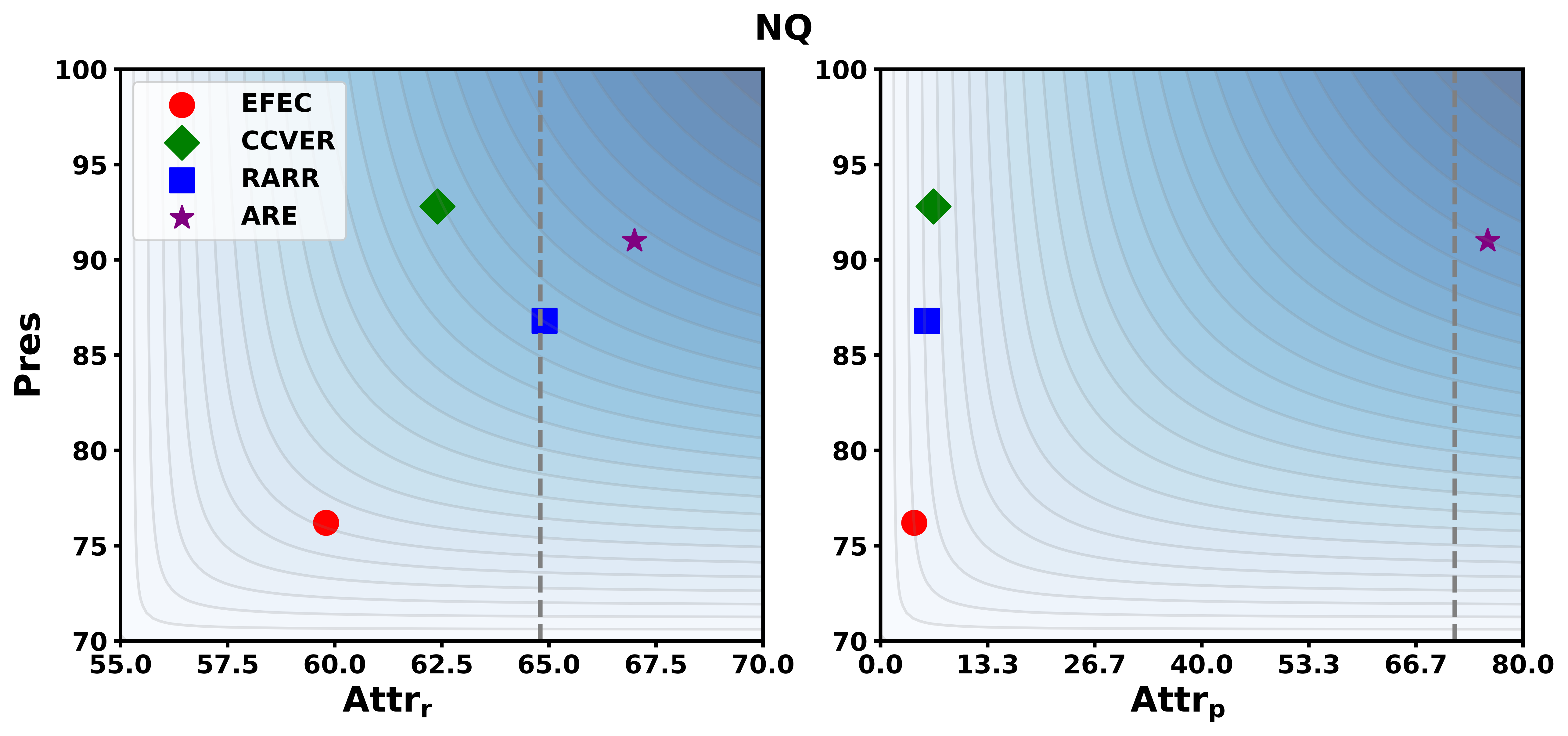}
    \end{minipage}  
    \begin{minipage}{\linewidth}
        \centering
        \includegraphics[scale=0.4]{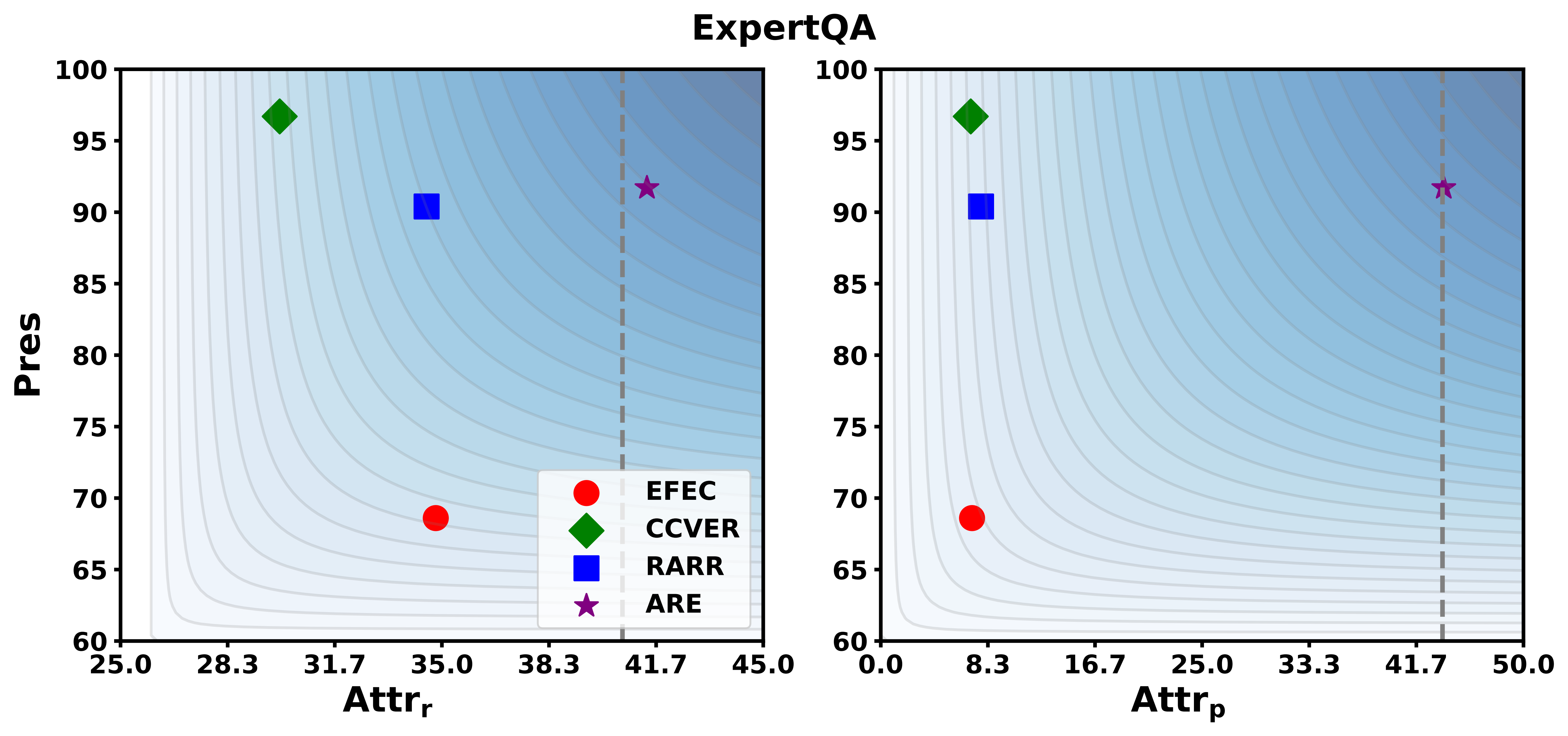}
    \end{minipage}
    \caption{The visualization experiments on NQ and ExpertQA datasets. The dashed line represents the highest attribution scores of all methods before editing. The points represent the performance of various methods after editing.
    The contours display level curves for $\mathrm{F1_{RP}}$ (left) and $\mathrm{F1_{PP}}$ (right), where points on the same contour share the same value. The closer a point is to the upper right corner, the better its performance is represented. 
    }
    \label{vis1}
\end{figure}

\subsection{Experimental Results and Analysis}

The experimental results are displayed in Table \ref{tab:main1} and Table \ref{tab:main2}, and the experimental analyses are listed as follows:

(1) ARE demonstrates substantial improvements in both molecular-level attribution evaluation metrics and intent preservation across four datasets using various LLMs. For instance, on Mintaka with API-based LLMs, ARE improves Ave-$\mathrm{F1_{PP}}$ by 72.4\% and Ave-$\mathrm{F1_{RP}}$ by 1.6\% (Table \ref{tab:main1}). With open-source LLMs (Table \ref{tab:main2}), ARE also achieves state-of-the-art performance—for example, in NQ, Ave-$\mathrm{F1_{PP}}$ and Ave-$\mathrm{F1_{RP}}$ improve by 9.5\% and 68.3\%, respectively—demonstrating the strong effectiveness and generalizability of our approach. The relatively smaller improvements observed on ExpertQA can be attributed to: (1) its specialized questions that often require expert-level domain knowledge beyond standard open knowledge bases, and (2) its significantly longer answers, averaging over 20 tokens more than other datasets.

(2) For attribution ability, ARE outperforms other baselines in $Attr_r$ and $Attr_p$. Especially in $Attr_p$, all existing methods are far below ARE. This is because CCVER and RARR retrieve a larger quantity of evidence, which is typically fragmented. Although fragmented evidence can be beneficial for $Attr_r$ due to its focus on entailment probability, $Attr_p$ is a stricter metric that evaluates the completeness of the evidence. These evidence cannot support the entire sentence, resulting in poor performance in $Attr_p$. As for \textit{Pres}, although the CCVER gets a higher \textit{Pres} in GPT-3.5 and Llama3-70B, it performed poorly on Llama2-70B. In contrast, ARE achieved the most stable \textit{Pres} across all LLMs, even with the less effective Llama2-70B, particularly when using the RARR method. These results demonstrate that ARE can maximize the attribution score while preserving the original intent.

(3) The sub-question generation methods perform similarly on NQ, Mintaka and StrategyQA, but CCVER significantly underperforms compared to RARR on ExpertQA. This highlights the limitations of sub-question generation, which requires carefully designed prompts. Conversely, ARE shows greater robustness and adaptability across different datasets.

\subsection{Ablation Study}
In order to investigate the impact of key modules on experimental performance, we conduct a series of ablation experiments, and the corresponding results are presented in Table \ref{tab:ablation}. Specifically, ``\textit{w/o} edit'' means removing the editing module; ``\textit{w/o} atomic'' represents removing atomic-level facts; ``\textit{w/o} molecular'' represents removing molecular-level clause;  ``\textit{w/o} re-retrieval'' means removing the status of ``irrelevant'' in evidence verifier.

(1) In the ``\textit{w/o} edit'' setting, attribution scores decline significantly. For example, $Attr_r$ decreased by 4.2\%, while $Attr_p$ dropped by 2.2\% on NQ. These results underscore the importance of atomic fact editing, as modifying hallucinated content is crucial for further improving the accuracy of attribution.

(2) The results of the ``\textit{w/o} atomic'' setting show that atomic facts play an important role in the attribution scores. Specifically, ARE obtains improvements of 23. 9\% in $Attr_r$ and 11. 8\% in $Attr_p$ on NQ dataset. This phenomenon indicates that atomic facts have higher certainty and lower entropy, making it possible to retrieve relevant evidence more effectively.

(3) Although the ``\textit{w/o} molecular'' setting performs better on $Attr_r$ and $Attr_p$, it shows poorer performance on the \textit{Pres} metric (decreased by 32.5\% on NQ, 30.1\% on StrategyQA, 35.6\% on ExpertQA). This may be attributed to the removal of molecular facts, which causes atomic facts to lose their correspondence with them, preventing accurate backtracking to their original positions for editing. Consequently, this significantly increases the risk of altering the original intent.

(4) The performance of ``\textit{w/o} re-retrieval" shows that the necessity of the ``irrelevant'' status in evidence verifier module. For example, $Attr_r$ decreased by 8.1\% and 6\% in $Attr_p$ in the NQ dataset. This phenomenon demonstrates that expanding facts when they have irrelevant evidence and then re-retrieving evidence can effectively enhance the attribution capability.

\begin{table}[H]
\centering
\caption{Comparison with two prompt-based variants of ARE on Mintaka dataset using GPT-3.5.}
\label{tab:effectiveness}
\setlength{\tabcolsep}{0.15cm}{
\begin{tabular}{cccc:cc}
\toprule
\multicolumn{1}{c}{Methods} & $Attr_r$ & $Attr_p$ & Pres & $\mathrm{F1_{PP}}$ & $\mathrm{F1_{RP}}$ \\ \midrule
CCVER & 0.630 & 0.069 & 0.937 & 0.129 & 0.753 \\
RARR & 0.646 & 0.06 & 0.850 & 0.112 & 0.734 \\
Prompt-based two-stage ARE & 0.688 & 0.77 & 0.858 & 0.730 & 0.764 \\
Prompt-based ARE & 0.682 & 0.768 & 0.774 & 0.723 & 0.725 \\
\textbf{ARE} & 0.716 & 0.807 & 0.914 & \textbf{0.857} & \textbf{0.803} \\ \bottomrule
\end{tabular}%
}
\end{table}

\subsection{Visualization Results of Attribution Scores and Preservation}
To provide a more intuitive comparison between existing methods and our proposed ARE, regarding the simultaneous achievement of maximum attribution scores and preservation of original intent, we conducted the visualization experiments illustrated in Figure \ref{vis1}. ARE is located on the right side of the dashed line, indicating that ARE has achieved effective editing. 
The contour lines indicate that the proposed ARE achieves the best performance in both the F1$_\mathrm{RP}$ (F1$_\mathrm{RP}$ both considers Attr$_r$ and \textit{Pres}) and F1$_\mathrm{PP}$ (F1$_\mathrm{RP}$ simultaneously considers Attr$_p$ and \textit{Pres}) metrics after editing. These phenomena demonstrate the robustness of the proposed ARE in maximizing the attribution scores and preserving the original intent.

\subsection{Effectiveness of Instruction-tuning}

To demonstrate the effectiveness of our proposed instruction-tuned fact decomposition LLM, we further present two variants of ARE: 1) Prompt-based two-stage ARE: This method first prompts LLMs to generate molecular clauses and then uses them to generate atomic facts using prompts. 2) Prompt-based ARE: This method only replaces the instruction-tuned fact decomposition LLM with the prompt.

As illustrated in Table \ref{tab:effectiveness}, ARE significantly outperforms two variants of ARE, the performance decreased by about 12\% and 4\% in F1$_\mathrm{PP}$ and F1$_\mathrm{RP}$, respectively. Prompt-based ARE fails to capture the complex structure of long-form answers, and prompt-based two-stage ARE still performs less than ARE, although it simplifies the process of generating atomic facts by first generating molecular clauses. Furthermore, ARE gives better performance because it utilizes a fact decomposition LLM tuned on a well-constructed dataset that enables ARE to better capture the complex structure of long-form answers.

\subsection{Experiments of Aggregating Evidence from Sub-questions}
The sub-question generation method (such as RARR) lacks a clear correspondence between questions and clauses, thus failing to provide complete evidence for molecular clauses. To demonstrate this, we employ our Molecular-to-Atomic fact decomposition LLM to first decompose the long-form answer into molecular clauses and then generate sub-questions for every molecular clause using GPT-3.5. Finally, the evidences retrieved through the sub-questions are aggregated according to the molecular clause. As shown in Table \ref{tab:my-table111}, the performance in $Attr_p$ is obviously improved (from 0.109 to 0.766 on NQ) by aggregating the evidence according to the molecular clauses (the method of ``RARR \textit{w/} aggregate''), showing that $Attr_p$ can evaluate the completeness. However, $Attr_r$ decreases, indicating that RARR retrieves irrelevant evidence that makes the aggregated evidence noisier.

\begin{table*}[ht]
\centering
\caption{The experimental results of aggregating evidence retrieved from sub-questions into the corresponding molecular clauses.}
\label{tab:my-table111}
\setlength{\tabcolsep}{0.16cm}{
\begin{tabular}{cccccc:ccccc:ccccc}
\toprule
\multirow{2}{*}{\textbf{Method}}  & \multicolumn{5}{c}{\textbf{NQ}} & \multicolumn{5}{c}{\textbf{StrategyQA}} & \multicolumn{5}{c}{\textbf{ExpertQA}} \\
\cline{2-16}
 & $Attr_r$ & $Attr_p$ & Pres & $\mathrm{F1_{PP}}$ & $\mathrm{F1_{RP}}$ & $Attr_r$ & $Attr_p$ & Pres & $\mathrm{F1_{PP}}$ & $\mathrm{F1_{RP}}$ & $Attr_r$ & $Attr_p$ & Pres & $\mathrm{F1_{PP}}$ & $\mathrm{F1_{RP}}$ \\ \hline
RARR & 0.649 & 0.058 & 0.868 & 0.109 & 0.743 & 0.356 & 0.097 & 0.862 & 0.174 & 0.504 & 0.34 & 0.078 & 0.904 & 0.144 & 0.494 \\
RARR $w/$ aggregate & 0.499 & 0.646 & 0.942 & 0.766 & 0.653 & 0.336 & 0.53 & 0.948 & 0.680 & 0.496 & 0.276 & 0.407 & 0.959 & 0.571 & 0.428 \\
\textbf{ARE} & 0.67 & 0.756 & 0.91 & \textbf{0.826} & \textbf{0.772} & 0.463 & 0.559 & 0.899 & \textbf{0.689} & \textbf{0.611} & 0.412 & 0.438 & 0.917 & \textbf{0.593} & \textbf{0.569} \\ \bottomrule
\end{tabular}%
}
\end{table*}

\subsection{The Performance of Fact Decomposition and Quality Analysis of Constructed Dataset}
\label{result}

\begin{figure}[t]
    \centering
       \includegraphics[scale=0.5]{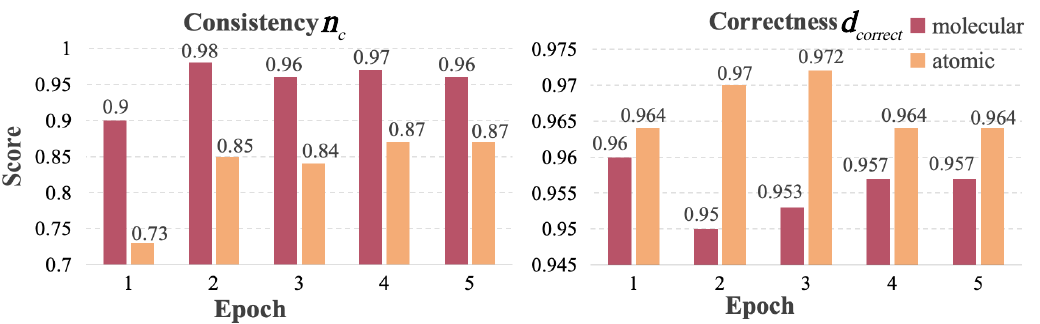}
    \caption{The performance of the fact decomposition LLM at the molecular level and the atomic level for different number of iterations.}
    \label{fig:fd}
\end{figure}

To find the best performance of molecular-to-atomic fact decomposition LLM and explore how iterations affect fact decomposition performance, we perform a comparative experiment, and the results are illustrated in Figure \ref{fig:fd}. Specifically, we assess its performance at both the molecular and atomic levels, focusing on two key aspects: consistency and correctness. Consistency $n_c$ measures how closely the number of decomposed sentences aligns with that of gold sentences. Correctness $d_{correct}$ evaluates whether the decomposed sentences preserve the original sentence meaning.

We test on a well-constructed evaluation dataset mentioned in \ref{section1} to select the decomposition model used in ARE. As shown in Figure \ref{fig:fd}, the performance of $n_c$ at the molecular-level initially increases and then decreases as the number of epochs increases. At the atomic-level, $n_c$ shows an overall upward trend, stabilizing at epoch 4. For $d_{correct}$, performance at molecular-level also increases and then decreases as the number of epochs increases, peaking at epoch 3 with 0.972. However, at the atomic-level, it first decreases and then increases. 

In terms of consistency, the performance at the 3rd and 4th epochs is similar. However, for correctness, the score of the atomic-level at epoch 3 is the highest, and meanwhile the score of the molecular-level at epoch 3 and epoch 4 has little difference. Therefore, we select the LLM trained with 3 epochs.

Although the fact decomposition LLM has achieved satisfactory performance, misaligned decomposition scenarios still exist. Since ARE is a multi-stage pipeline architecture, such misaligned decomposition may lead to error propagation. This issue represents one of the primary challenges in our system. Nevertheless, our multi-stage pipeline design still demonstrates substantial overall gains, and the benefits significantly outweigh the drawbacks introduced by occasional decomposition errors. As shown in Table \ref{tab:ablation}, the performance has decreased when removing the editing (``\textit{w/o} edit'') or re-retrieval (``\textit{w/o} re-retrieval'') stage respectively, which demonstrates that ARE can address the error propagation. 

\begin{table}[h]
    \centering
    \caption{The quality report of the generated instruct-tuning dataset}
    \setlength{\tabcolsep}{0.8cm}{
    \begin{tabular}{ccc}
        \toprule
        Dataset & Human & LLM-as-a-Judge\\ \midrule
        Train (500) & 0.934 & 0.965\\
        Test (200) & 0.946 & 0.972\\
        average & 0.940 & 0.969\\
        \bottomrule
    \end{tabular}
    
    \label{quality report}
}
\end{table}

We evaluated the quality of the generated instruct-tuning dataset (randomly selected 500 and 200 from train and test respectively) using both human evaluation and LLM-as-a-Judge (Llama-3.1-70B-Instruct). The accuracy scores for LLM generated long-form text, molecular clauses, and atomic facts were 0.94 (human) and 0.969 (LLM), as shown in Table \ref{quality report}.

\begin{table}[t]
\centering
\caption{Comparison of the proposed atomic fact-based retrieval with previous sub-question-based retrieval performance.
}

\label{tab:only-retrieval}
\setlength{\tabcolsep}{0.23cm}{
\begin{tabular}{ccccccc}
\toprule
\multirow{2}{*}{\textbf{Methods}} & \multicolumn{2}{c}{NQ} & \multicolumn{2}{c}{StrategyQA} & \multicolumn{2}{c}{ExpertQA} \\
\cline{2-7}
 & $Attr_r$ & $Attr_p$ & $Attr_r$ & $Attr_p$ & $Attr_r$ & $Attr_p$ \\ \midrule
DRQA & 0.424 & 0.647 & 0.237 & 0.49 & 0.127 & 0.283 \\
CCVER & 0.602 & 0.044 & 0.359 & 0.044 & 0.296 & 0.073 \\
RARR & 0.593 & 0.043 & 0.302 & 0.096 & 0.305 & 0.077 \\
\textbf{ARE} & \textbf{0.648} & \textbf{0.715} & \textbf{0.442} & \textbf{0.550} & \textbf{0.404} & \textbf{0.437} \\ \bottomrule
\end{tabular}%
}
\end{table}

\begin{table}[t]
\centering
\caption{Comparison of inference efficiency in latency, interaction times, token consumed and retrieval times. For a fair compare, we use Llama3-70B as the base LLM, running on 4$\times$A800 GPUs to eliminate network fluctuation influence.}
\label{tab:Efficiency}
\setlength{\tabcolsep}{0.65cm}{
\begin{tabular}{lcc}
\toprule
Methods & \makecell{\textbf{Latency} $\downarrow$ \\ (sample/s)} & \makecell{\textbf{Interaction} $\downarrow$ \\ (times/sample)} \\ \midrule
RARR & 55 & 7.1 \\
CCVER & 87 & 9 \\
\textbf{ARE} & \textbf{41} & \textbf{4.7} \\ \cline{1-3} 
& \makecell{\textbf{Consumed tokens} $\downarrow$ \\ (tokens/sample)} & \makecell{\textbf{Retrieval} $\downarrow$ \\ (times/sample)} \\ \cline{1-3}
RARR & 7371 & 5.93 \\
CCVER & 9711 & 7 \\
\textbf{ARE} & \textbf{5340} & \textbf{4.77} \\ \bottomrule
\end{tabular}%
}
\end{table}

\subsection{Hypothesis Proof Experiment}
\label{hyposis}
As illustrated in Table \ref{tab:only-retrieval}, ARE shows the improvements of 4.6\% on NQ, 8.3\% on StrategyQA, and 9.9\% on ExpertQA based on the metric of $Attr_r$. This supports the hypothesis that atomic facts are easier to retrieve as relevant evidence. Furthermore, DRQA, using the entire answer and question for retrieval, struggles in cross-document retrieval scenarios. Methods involving generated sub-questions fail to obtain more evidence due to the poor quality of generated questions.

CCVER and RARR both generate sub-questions and then use them for retrieval. Their performance is similar on NQ and ExpertQA, but on StrategyQA, CCVER outperforms RARR. This is because CCVER generates ``Yes or No'' type questions, which match the type of the StrategyQA dataset, allowing for better evidence retrieval. However, both perform worse than ARE, which is based on atomic fact retrieval.

\subsection{Inference Efficiency Analysis}

To compare the inference efficiency between ARE and other baselines, we perform controlled inferences CCVER, RARR and ARE on 4 NVIDIA A800 GPUs using VLLM\footnote{https://github.com/vllm-project/vllm}. We measure average latency, interaction times with LLMs, and token consumption. As shown in Table \ref{tab:Efficiency}, ARE achieves a shorter average inference time per sample compared to RARR. The average interaction times with the LLMs are 4.7, and the average tokens consumed are 5340, both of which outperform the baselines. This is because the number of atomic facts generated by instruction-tuned decomposition LLM is less, but they are more relevant to the original answer than RARR and CCVER, leading to fewer retrievals (\textbf{4.77} times for ARE, 5.93 for RARR and 7 for CCVER).

\begin{table}[ht]
    \centering
    \caption{Comparing the revised results with different systems. \textcolor{red}{Red} indicates hallucinated content that conflicts from the evidence or fact. \textcolor{blue}{Blue} marks the content that, after revision, has been corrected to align with the evidence. \textcolor{orange}{Orange} indicates it is still not correct after the revision.}
    \label{tab:case1}
    \begin{tabularx}{\linewidth}{X}  
    \toprule
    \multicolumn{1}{c}{\textbf{Original long-form answer}} \\ 
    "Daddy's Little Girl" is a song by American \textcolor{red}{country music artist Kellie Pickler}. The song was released in 2006 as the second single from her debut album "Small Town Girl." 
    
    \textbf{Gold Answer}: ``Al Martino'' \\ \midrule
    
    \multicolumn{1}{c}{\textbf{Revised by EFEC}} \\ 
    "The Daddy's Little \textcolor{orange}{Girl is a book written and recorded by a Canadian music producer, and author."} \ \ \ \ \ \ \textbf{Pres: 0.322}\\ \midrule
    \multicolumn{1}{c}{\textbf{Revised by RARR}} \\ 
    "Daddy's Little Girl" is a \textcolor{orange}{classic song typically played at white weddings while a bride dances with her father. The song's lyrics and music were first written by Robert Harrison Burke and Horace Gerlach in 1949.} \ \ \ \ \ \ \textbf{Pres: 0.245}\\ \midrule
    \multicolumn{1}{c}{\textbf{Revised by ARE}} \\ 
    "Daddy's Little Girl" is a song by American \textcolor{blue}{traditional pop and jazz singer Al Martino}. The song was released in 2006 as the second single from her debut album "Small Town Girl." \ \ \ \ \ \ \textbf{Pres: 0.83} \\
    \bottomrule
       \end{tabularx}

\end{table}

\begin{table}[h]
    \centering
    \setlength{\tabcolsep}{0.3em}
    \caption{Error case in ARE. \textcolor{red}{Red} indicates the hallucinated content. \textcolor{blue}{Blue} marks as the correct content. \textcolor{orange}{Orange} represents the invalid evidence. \textcolor{green}{Green} is the specialized word.} 
    \begin{tabular}{p{0.8cm}p{1.5cm}p{6cm} }
    \toprule
         & Atomic Fact & The Euro is the official currency of Spain and is used by \textcolor{red}{19} of the 27 European Union countries.\\ \cline{2-3}
       \multirow{-2}{*}{case 1} &  Retrieved evidence& ...The euro is the official currency of \textcolor{blue}{20} of the 27 member states of the European Union.\\ \cline{1-3}
       & Atomic Fact &  \textcolor{red}{Chile} is one of the countries with the highest forest cover in the world.\\ \cline{2-3}
       \multirow{-2}{*}{case 2}& Retrieved evidence & This is far more than the second greatest country in terms of forest area coverage, which is  \textcolor{blue}{Russia} . Russia has approximately ....  \\ \cline{1-3}
         & Long-form answer (Mintaka) & The first book in the Hunger Games series, titled ``\textcolor{blue}{The Hunger Games},'' was published in 2008. (Total tokens:  16)\\ \cline{2-3}
         & Atomic Fact & \textcolor{red}{The first book} was published in 2008.\\ \cline{2-3}
       \multirow{-3}{*}{case 3} &  Retrieved evidence& \textcolor{orange}{The Center for Fiction The Center for Fiction 2008 First Novel Prize .... This annual award was created in 2006 to honor the best first novel of the year.}\\ \cline{1-3}
       & Long-form answer (ExpertQA) &  A client with substance addiction, .... Secondly, \textcolor{green}{Dialectical Behavior} .... Additionally, \textcolor{green}{Cognitive-Behavioral Therapy (CBT)} .... Finally, a \textcolor{green}{trauma-informed} approach to addiction.... (Total tokens:  111)\\ \cline{2-3}
       & Atomic Fact &  A client with substance addiction, ... requires a comprehensive and integrative therapeutic approach.\\ \cline{2-3}
       \multirow{-2}{*}{case 4}& Retrieved evidence & \textcolor{orange}{... of childhood trauma in a range of settings that are particularly likely to serve this population like addictions, ... in the context of the past trauma.}  \\ \bottomrule
    \end{tabular}
    
    \label{tab:my_label_case}
\end{table}

\subsection{Case Study}
When comparing the revised results of EFEC, RARR, and ARE in Table \ref{tab:case1}, it clearly highlights its advantages, as well as the shortcomings of RARR and EFEC. ARE performs minimal edits by focusing on atomic facts, successfully preserving the original intent while improving factuality and attribution.  In contrast, RARR often makes substantial modifications, which can make it challenging to preserve the original intent of the content. EFEC  performs even less satisfactorily. It makes significant changes to the original content, omitting entire sentences. Such drastic revisions can result in a final product that significantly deviates from the original content and intent.

As shown in Table \ref{tab:my_label_case}, cases 1 and 2 illustrate verification errors where fine-grained differences between evidence and facts are missed, leading to incorrect classifications (e.g., labeling ``editing required'' as ``supportive'' or vice versa). We randomly selected 50 samples for manual evaluation, finding a 23.1\% error rate in misclassifying ``editing required'' as ``irrelevant''. In case 1, the evidence cites 20 Eurozone countries while the fact mentions 19, but this discrepancy is not detected. Case 2 presents irrelevant evidence about Russia’s forests for a fact concerning Chile.
Case 3 shows a decomposition failure where the LLM fails to extract the key subject “The Hunger Games” from a long-form answer, resulting in an invalid atomic fact. Case 4 highlights compounded failure in medical-domain content involving a long (111-token) multi-fact structure and inductive reasoning. The decomposition is insufficient, leading to ineffective retrieval and cascading errors across subsequent stages.

\begin{table}[h]
\centering
\caption{The prompt templates used in ARE.}
\label{tab:prompt}
\setlength{\tabcolsep}{0.2cm}{
\begin{tabular}{cl}
\toprule
\textbf{LLMs} & \makecell{\textbf{Prompts}} \\ \hline
\begin{tabular}[c]{@{}c@{}}Long-form 
\\ answer \\ generation\end{tabular} & \begin{tabular}[c]{@{}l@{}}You need to think step by step and answer my question. \\ Example 1: \\ 1.Question: \textbackslash n 2.Explanation: \textbackslash n 3.Answer:  \\ Please answer the question and give the Explanation and \\ an answer.\end{tabular} \\ \midrule
\begin{tabular}[c]{@{}c@{}}Evidence 
\\ verifying\end{tabular} & \begin{tabular}[c]{@{}l@{}}I will check some things you said. Here are some \\ examples for you to learn this process: \\ Example 1: \\ 1. You said: \textbackslash n 2. I checked: \textbackslash n 3. I found this article: \\ 4. Reasoning: \textbackslash n 5. Therefore: \\ Now, please follow the above examples to inference.\end{tabular} \\ \midrule
\begin{tabular}[c]{@{}c@{}}Fact 
\\ expansion\end{tabular} & \begin{tabular}[c]{@{}l@{}}Rewrite the fact into two short atomic phrases based on \\ Wikipedia. Ensure they contain more factual information \\ and can be easily supported by search engines.\\\end{tabular} \\ \midrule
Editing & \begin{tabular}[c]{@{}l@{}}You need use some evidences to check 'You said' if \\ there are some difference and generate in "My fix". \\ Here are some examples to learn: \\ Example 1: \textbackslash n  1. You said: \textbackslash n  2. I found these \\ evidences:  3.  This suggests \textbackslash n  4. My fix: \\I will give you a new instance, please follow the above \\ example  to fix a new one and start with "My fix:". \\\end{tabular} \\ \midrule
Backtracking & \begin{tabular}[c]{@{}l@{}}You are a linguist. Your task is to compose coherent\\ sentences using the provided atomic facts. While doing \\ so,  refer to the structure of the given reference text — \\ including sentence count, subject order, and pronoun \\ usage —  to maintain consistency and fluency.
\\ 
Here are some examples: \\

\{
``reference answer'': long-form answer \\

[\{Molecular clause 1: [Atomic facts]\}, \\

\{Molecular clause 2: [Atomic facts]\}, \\ \{ Molecular clause 3: [Atomic facts] \}]\} \\

.... \\
Now, think step by step to merge the given  facts by \\ referring to the reference text.   Generate the output in \\ coherent sentences starting with "Sentences:".

\\\end{tabular} \\

\bottomrule
\end{tabular}%
}
\end{table}

\section{Conclusion}
This paper proposes a novel \textbf{A}tomic fact decomposition-based \textbf{R}etrieval and \textbf{E}diting (\textbf{ARE}) framework for post-hoc retrieval-based AQA tasks, which contains a Molecular-to-Atomic decomposition stage and an Atomic-to-Molecular backtracking process. Specifically, ARE employs an instruction-tuned fact decomposition LLM to decompose the long-form answers into multiple molecular clauses and atomic facts. This LLM is fine-tuned in a well-constructed molecular-to-atomic fact decomposition dataset. Subsequently, ARE leverages an LLM-based verifier to validate the relationships between the searched evidences and their corresponding atomic facts. Based on the verifier's assessment, 
ARE determines whether the facts require  further expansion for re-retrieval or editing. Furthermore, ARE proposes a more comprehensive evaluation metric $Attr_p$, which not only accurately measures the precision of retrieved evidence, but also emphasizes the completeness of the evidence. The effectiveness of this framework is demonstrated across four datasets using four prominent LLMs, complemented by an extensive ablation study and LLM evaluation.

\section{ACKNOWLEDGMENT}
This research was supported by NSFC (No.62376144), by Science and Technology Cooperation and Exchange Special Project of Shanxi Province (No.202204041101016), by Key Research and Development Project of Shanxi Province (No.202102020101008), by Natural Language Processing Innovation Team (Sanjin Talents) Project of Shanxi Province. Jiaoyan is funded by the EPSRC project OntoEm (EP/Y017706/1)

\bibliographystyle{IEEEtran}
\bibliography{ARE}

\begin{IEEEbiography}[{\includegraphics[width=1in,height=1.25in,clip,keepaspectratio]{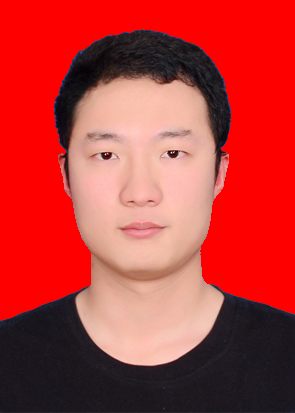}}]
{Zhichao Yan} is currently working toward the PhD degree with the School of Computer and Information Technology, Shanxi University. His research interests include question answering, large language models and information retrieval.
\end{IEEEbiography}
\begin{IEEEbiography}[{\includegraphics[width=1in,height=1.25in,clip,keepaspectratio]{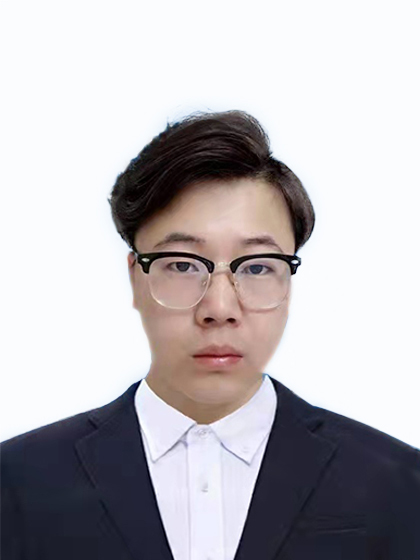}}]
{Jiapu Wang} is currently a Ph.D student in the Beijing Municipal Key Laboratory of Multimedia and Intelligent Software Technology, Beijing University of Technology, Beijing. His research interests include knowledge graph completion, large language model reasoning.
\end{IEEEbiography}
\begin{IEEEbiography}[{\includegraphics[width=1in,height=1.25in,clip,keepaspectratio]{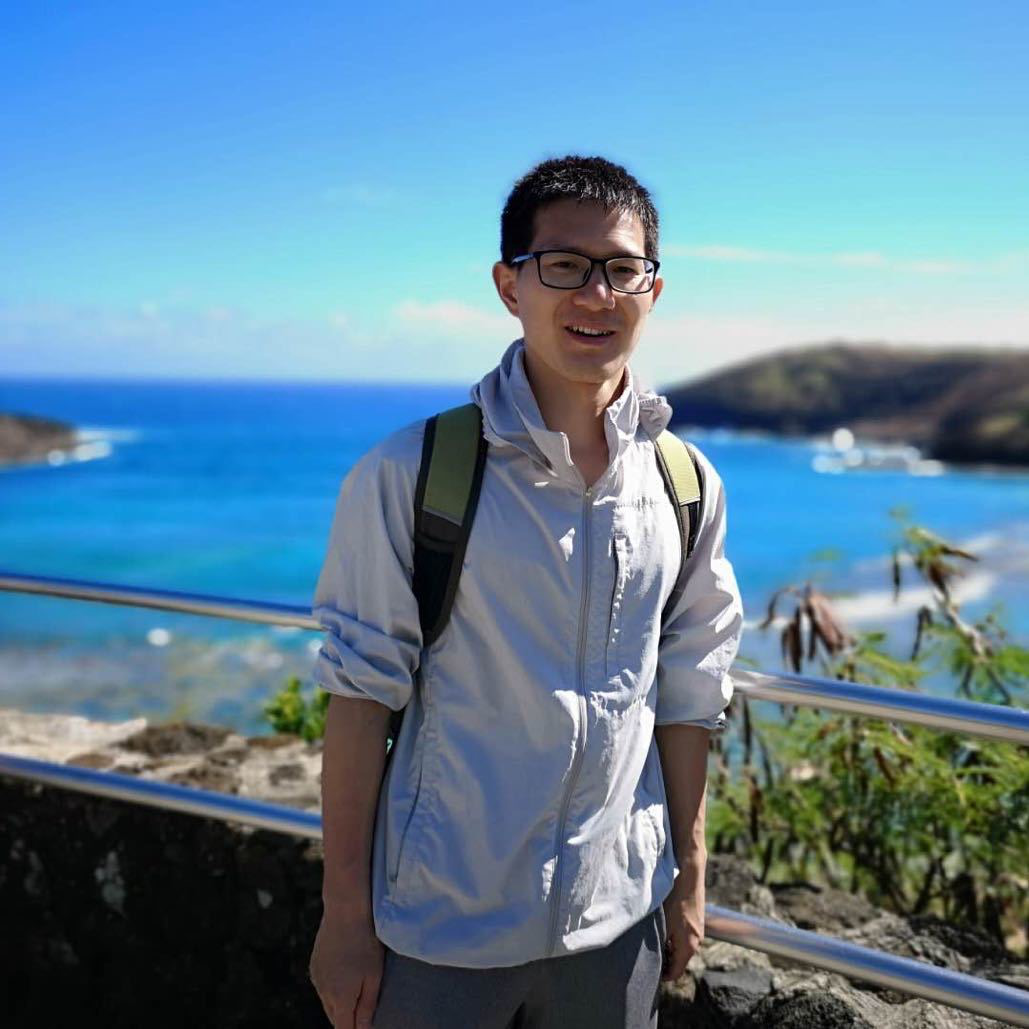}}]
{Jiaoyan Chen} is a Lecturer in Department of Computer Science, The University of Manchester, and a part-time senior researcher in Department of Computer Science, University of Oxford. His research interests mainly lie in Knowledge Representation, Knowledge Graph, Ontology, and Machine Learning, with over 10 years research experience and over 60 papers published in top Computer Science conferences and journals including AAAI, IJCAI, WWW, ISWC, ICLR, PIEEE, Machine Learning, JWS and so on. 
\end{IEEEbiography}

\begin{IEEEbiography}[{\includegraphics[width=1in,height=1.25in,clip,keepaspectratio]{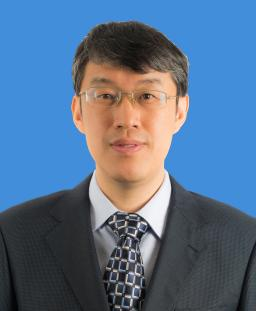}}]
{Xiaoli Li} (Fellow, IEEE) is Professor of Singapore University of Technology and Design, and adjunct Professor, School of Computer Science and Engineering, Nanyang Technological University. His research interests include data mining, machine learning, artificial intelligence and bioinformatics. Regional Area Chair of data mining and artificial intelligence conferences, Senior Program Committee member/Workshop Chair, Editor-in-Chief of the Journal of Artificial Intelligence Review, Associate Editor of IEEE Transactions on Artificial Intelligence, Knowledge and Information Systems, and Journal of Machine Learning and Applications. 
\end{IEEEbiography}

\begin{IEEEbiography}[{\includegraphics[width=1in,height=1.25in,clip,keepaspectratio]{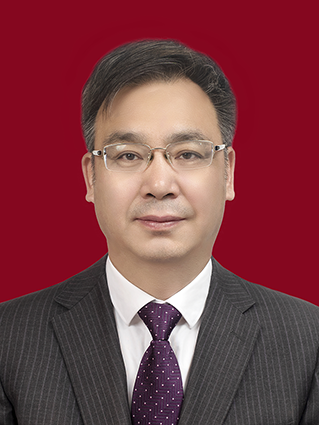}}]
{Jiye Liang} (Fellow, IEEE) received the Ph.D. degree from Xian Jiaotong University. He is a professor in Key Laboratory of Computational Intelligence and Chinese Information Processing of Ministry of Education, the School of Computer and Information Technology, Shanxi University. His research interests include artificial intelligence, granular computing, data mining, and machine learning. He has published more than 300 papers in his research fields, including AI, JMLR, IEEE TPAMI, IEEE TKDE, ML, NeurIPS, ICML, AAAI.
\end{IEEEbiography}

\begin{IEEEbiography}[{\includegraphics[width=1in,height=1.25in,clip,keepaspectratio]{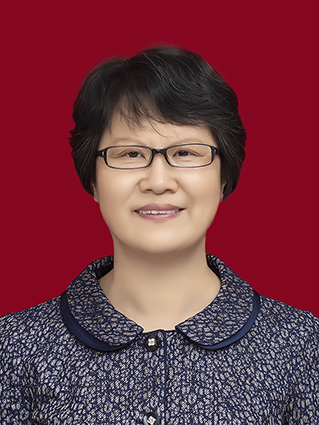}}]
{Ru Li} is currently a professor with the Key Laboratory of Computational Intelligence and Chinese Information Processing of Ministry of Education, School of Computer and Information Technology, Shanxi University. She received her Ph.D. degree in computer science from Shanxi University, China. She has published numerous papers in important academic journals and conferences, such as IEEE TKDE, ACL, IJCAI, EMNLP, CIKM, COLING, and so on. Her research interests include natural language processing, text semantics analysis, and machine learning.
\end{IEEEbiography}

\begin{IEEEbiography}[{\includegraphics[width=1in,height=1.25in,clip,keepaspectratio]{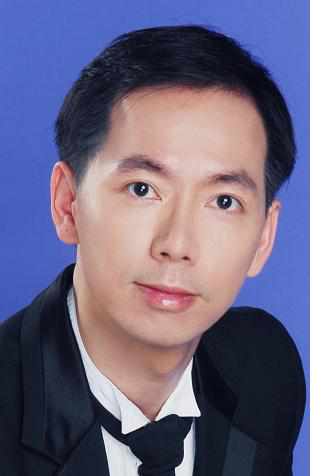}}]
{Jeff Z. Pan} is currently a lifetime professor at the University of Edinburgh, UK, and serves as the Chair of Knowledge Graphs at the Alan Turing Institute, Director of the Huawei Knowledge Graph Lab in Edinburgh, Huawei's Chief Search Scientist in the UK. He received his Ph.D. degree in computer science from the University of Manchester, UK in 2004. He has published over 200 papers in important academic journals and conferences, such as AIJ, IEEE TKDE, ACL, SIGIR, IJCAI, AAAI, and so on. He serves as the associate editor for top knowledge graph journals TGDKS and JWS. His research interests focus on knowledge representation, knowledge-based learning, and knowledge-based NLP.
\end{IEEEbiography}

\vfill

\end{document}